%% file: main.tex
\documentclass[sigconf]{acmart}
\input{preamble}

\usepackage{fancyhdr}

\begin{document}

\renewcommand{\headrulewidth}{0pt}

\fancyhf{}
\fancyhead[C]{Submitted to ACM for possible publication}

\thispagestyle{fancy}

\title[\FW: Sensing-Assisted Wireless Edge Computing]{\FW: Sensing-Assisted Wireless Edge Computing}


\author{Khandaker Foysal Haque\textsuperscript{\textdagger}, Francesca Meneghello\textsuperscript{*}, Md. Ebtidaul Karim\textsuperscript{\textdagger}, and Francesco Restuccia\textsuperscript{\textdagger}}
\affiliation{%
  \institution{\textsuperscript{\textdagger}Institute for the Wireless Internet of Things, Northeastern University, United States\\
  \textsuperscript{*}Department of Information Engineering, University of Padova, Italy}
\streetaddress{}
\city{}
\country{}}

\renewcommand{\shortauthors}{K.F. Haque, F. Meneghello, Md. Karim and F. Restuccia}
\vspace{2cm}

\begin{abstract}
\noindent Emerging mobile \gls{vr} systems will require to continuously perform complex computer vision tasks on ultra-high-resolution video frames through the execution of \glspl{dnn}-based algorithms. Since state-of-the-art \glspl{dnn} require computational power that is excessive for mobile devices, techniques based on \gls{wec} have been recently proposed. However, existing \gls{wec} methods require the transmission and processing of a high amount of video data which may ultimately saturate the wireless link. In this paper, we propose a novel \textit{Sensing-Assisted Wireless Edge Computing} (\FW) paradigm to address this issue. \FW leverages knowledge about the physical environment to reduce the end-to-end latency and overall computational burden by transmitting to the edge server only the relevant data for the delivery of the service. Our intuition is that the transmission of the portion of the video frames where there are no changes with respect to previous frames can be avoided. Specifically, we leverage wireless sensing techniques to estimate the location of objects in the environment and obtain insights about the environment dynamics. Hence, only the part of the frames where any environmental change is detected is transmitted and processed. We evaluated \FW by using a 10K \cam camera with a Wi-Fi 6 sensing system operating at 160~MHz and performing localization and tracking. We considered instance segmentation and object detection as benchmarking tasks for performance evaluation. We carried out experiments in an anechoic chamber and an entrance hall with two human subjects in six different setups. Experimental results show that \FW reduces both the channel occupation and end-to-end latency by more than 90\% while improving the instance segmentation and object detection performance with respect to state-of-the-art \gls{wec} approaches. For reproducibility purposes, we pledge to share our dataset and code repository.
\end{abstract}

\maketitle

\glsresetall
\section{Introduction}

\pagestyle{fancy}
\thispagestyle{fancy}
Emerging technologies based on mobile \gls{vr}, such as the Metaverse, will provide new entertainment applications \cite{MusicMetaverse,SportsMetaverse}, providing ultra-realistic online learning experiences \cite{kye2021educational}, and transforming healthcare through remote surgery opportunities~\cite{DoctorsMetaverse}. 

A key issue currently stymieing the Metaverse is that commercial VR headsets do not deliver adequate performance to the end user.
Experts believe that \cam~ video frames should have at least 120 Hz frame rate with 8K resolution to avoid pixelation and motion sickness \cite{somrak2019estimating, gallagher2018cybersickness, PixelMetaverse}. However, current wireless \gls{vr} headsets on the market provide up to 4K resolution, with only a limited few achieving a frame rate within the range of 100-120 Hz~\cite{Soni_2024, HTC-Vive-Focus-3}.

To further complicate matters, VR systems need to continuously execute \glspl{dnn} to perform object detection \cite{wu2020recent} and semantic segmentation \cite{mo2022review} to include physical expressions into digital avatars \cite{tao2019make}. Existing mobile VR headsets do not have enough computational resources to execute complex \gls{dnn} tasks like object detection and segmentation on 8K frames~\cite{somrak2019estimating}. Thus, they either excessively decrease battery lifetime \cite{OverheatingVRHeadset} or degrade the performance to unacceptable levels \cite{matsubara2022bottlefit}. While mobile \glspl{dnn} such as MobileNet~\cite{sandler2018mobilenetv2} and MnasNet~\cite{tan2019mnasnet} can decrease the computation requirements, they lose in accuracy -- up to 6.4\%  -- compared to large \glspl{dnn} such as ResNet-152~\cite{he2016deep}. While \gls{wec} can address the issue, continuously offloading \gls{dnn} tasks requires transmission data rates that far exceed what existing wireless technologies can offer today. Indeed, sending frames at 120 Hz with 8K resolution would require about 40 Gbps of data rate for each AR/VR device, while today, Wi-Fi supports a maximum of 1.2 Gbps network-wide \cite{ieee-80211ax, 8847238}.

\begin{figure}[t]
    \centering
    \includegraphics[width=\columnwidth ]{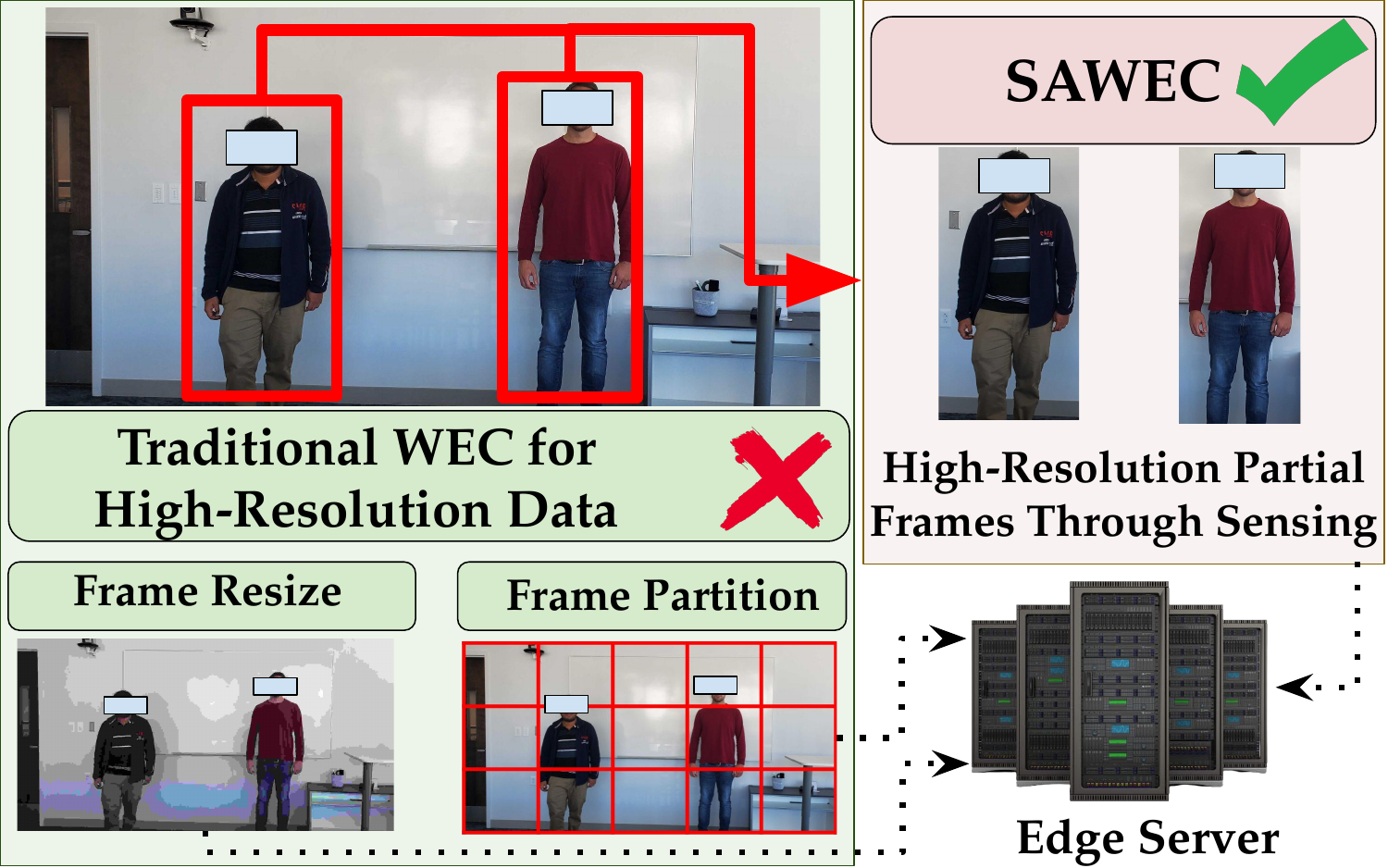}
    \setlength\abovecaptionskip{-0.1cm}
    \caption{\FW vs traditional edge computing approaches. \vspace{-0.2cm}}
    \label{fig:SAWEC overview}
\end{figure}

Another key issue is that existing \glspl{dnn} are not trained on 8K images -- for example, object detection and segmentation models like YOLOv8~\cite{Jocher_YOLO_by_Ultralytics_2023} and SAM~\cite{kirillov2023segment} are often benchmarked with COCO dataset \cite{lin2014microsoft} and SA-1B V1.0 dataset \cite{kirillov2023segment} which are respectively of size 640$\times$640 and 1500$\times$2250. While compressing and/or downsizing the frames would reduce the data rate requirement, \textit{we show through experiments in Section \ref{sec: background_and_moti} that downsizing by a factor smaller than 0.5 or employing a compression ratio of less than 0.125 would respectively reduce the system performance by 80\% and 60\%}. While partitioning the 8K frames into multiple smaller tiles can be another option, the region of interest might fall into multiple tiles causing performance degradation. \emph{We show in Section \ref{sec:Segmentation Performance} that partitioning causes up to 46.05\% of performance loss}.

To address the above challenges, we propose \textit{Sensing-Assisted Wireless Edge Computing }(\FW). As depicted in Figure \ref{fig:SAWEC overview}, \FW offloads to the edge server only the relevant portions of the frames -- thus decreasing data rate requirements with respect to traditional edge computing approaches, without losing any image resolution. In stark opposition with existing art that compresses or partitions frames~\cite{jiang2023energy, yang2022edgeduet, wang2023task}, \FW leverages wireless sensing to localize and track environmental changes such as the movements of humans or other objects. This way, instead of offloading the whole frame, \FW only offloads the part of the frame where motion is detected, hereafter referred to as \gls{roi}.  Since wireless sensing based localization can operate concurrently with wireless communications, \FW does not require any dedicated infrastructure as it leverages the \gls{cfr} computed through the channel estimation procedure which is routinely required by any wireless communication standard such as Wi-Fi (IEEE 802.11). Even though it would be possible to take similar approaches with event-based cameras~\cite{iaboni2021event} or software-only-vision~\cite{uhlrich2023opencap}, the approach in \FW is much simpler which also does not need any dedicated additional device. 

\smallskip
\noindent \textbf{\textit{Summary of Novel Contributions}}\smallskip

\noindent$\bullet$ We present \FW, a novel paradigm that performs wireless edge computing by leveraging wireless sensing integrated with the communication process (Section \ref{sec:system_overview}). \FW  optimizes the edge offloading process by localizing and tracking environmental changes that eventually determine the \gls{roi} to offload to the edge server instead of the whole frame. This way, \FW optimizes the transmission latency and channel utilization while improving the performance of the offloaded edge computing task; 

\noindent$\bullet$ A key challenge is that since existing IEEE 802.11ac (\mbox{Wi-Fi 5})-based Wi-Fi localization systems operate with up to 80 MHz bandwidth, they incur in relatively low range resolution. To have better localization granularity, in Section \ref{section: Wi-Fi-aided ROI Detection} we present a custom-tailored localization and tracking approach based on \gls{dbscan}, explained in Section \ref{subsec:clustering}. In addition, we propose a framework to project the estimated localization information in a \cam~video frame for detecting the \gls{roi} as explained in Section \ref{sec: adaptation};

\noindent$\bullet$ We implemented and validated \FW through several experiments in multiple propagation environments, i.e., an anechoic chamber and a hall room (Section \ref{sec:perf-ev}). In each of the environments, we performed an extensive data collection campaign with six different setups involving two human subjects (IRB approval is available upon request). In Section \ref{sec:exp-res}, we compare \FW with state-of-the-art work YolactACOS \cite{xie2022edge} and EdgeDuet \cite{yang2022edgeduet} and demonstrate that \FW reduces the channel occupation and end-to-end latency by more than 90\% while improving the instance segmentation and object detection $mAP_{50-95}$ performance by up to 45\%. {We share the whole dataset and code repository for reproducibility and research purposes at \textbf{\url{https://github.com/kfoysalhaque/SAWEC}}}.

\vspace{-0.1cm}
\section{Background and Related Work}\label{sec: background_and_moti}



Wireless edge computing (WEC) has gained significant traction over the last few years~\cite{islam2021survey, wang2020survey}. Prior work has focused on reducing the end-to-end latency as well as the channel occupation by employing \gls{dnn} partitioning \cite{ laskaridis2020spinn}, frame partitioning  \cite{yang2022edgeduet} and full task offloading \cite{xu2020approxdet}. While frame downsizing and frame compression approaches \cite{jiang2023energy} can effectively decrease energy consumption and channel usage, they can hardly be applied in \gls{vr} applications with 8K resolution and higher. 

\begin{figure}[t]
    \centering
    \includegraphics[width=\columnwidth ]{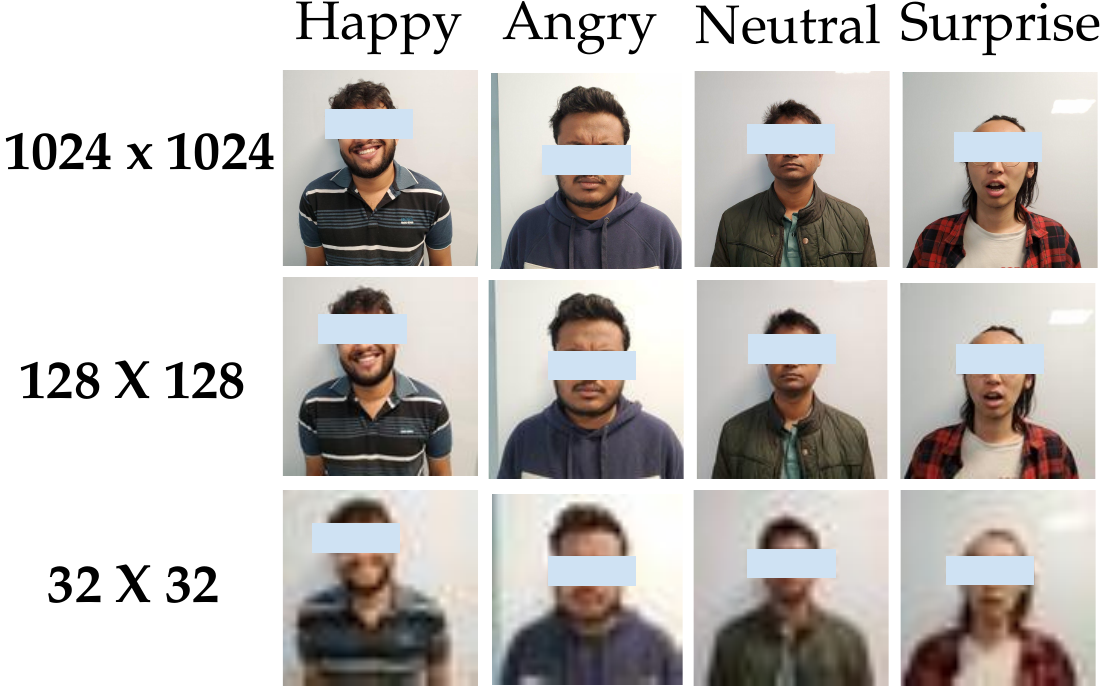}
    \caption{Samples from four different expressions of four subjects for image downsize ratio test.\vspace{-0.3cm}}
    \label{fig:expression}
\end{figure}

To show this point, we have performed an experiment on facial expression detection with different image downsize ratios ranging from 1 (image resolution 1024$\times$1024) to 1/16 (image resolution 64$\times$64) and image compression ratios ranging from 1 to 1/32. 
Image downsize refers to reducing the number of pixels used to represent an image. Whereas fewer pixels mean fewer data to store and offload the image, this process results in loss of details. On the other hand, image compression aims to make the images occupy less storage space without significantly compromising their quality, though some loss of detail is inevitable, especially at higher compression ratios. Specifically, we used JPEG compression in this work. 
We created a high-resolution facial emotion dataset with 1024$\times$1024-resolution images containing 8000 images from six subjects categorized into six facial emotion classes: \textit{angry, disgust happy, sad, surprise, and neutral}. Some examples are depicted in the first row of Figure \ref{fig:expression}. We implemented an 8-layer \gls{cnn} structure inspired by VGG-16 \cite{simonyan2014very} for facial expression detection. Hence, we compared the expression detection performance when using as input for the \gls{dnn} the original images or their downsized or compressed versions. The comparison has been performed considering the accuracy of the facial expression and the latency of the overall process, accounting for data compression, transmission and inference time.

\begin{figure}[t]
	\centering
	\subfloat [Image downsize ratio.]{%
		\includegraphics[width=\columnwidth]{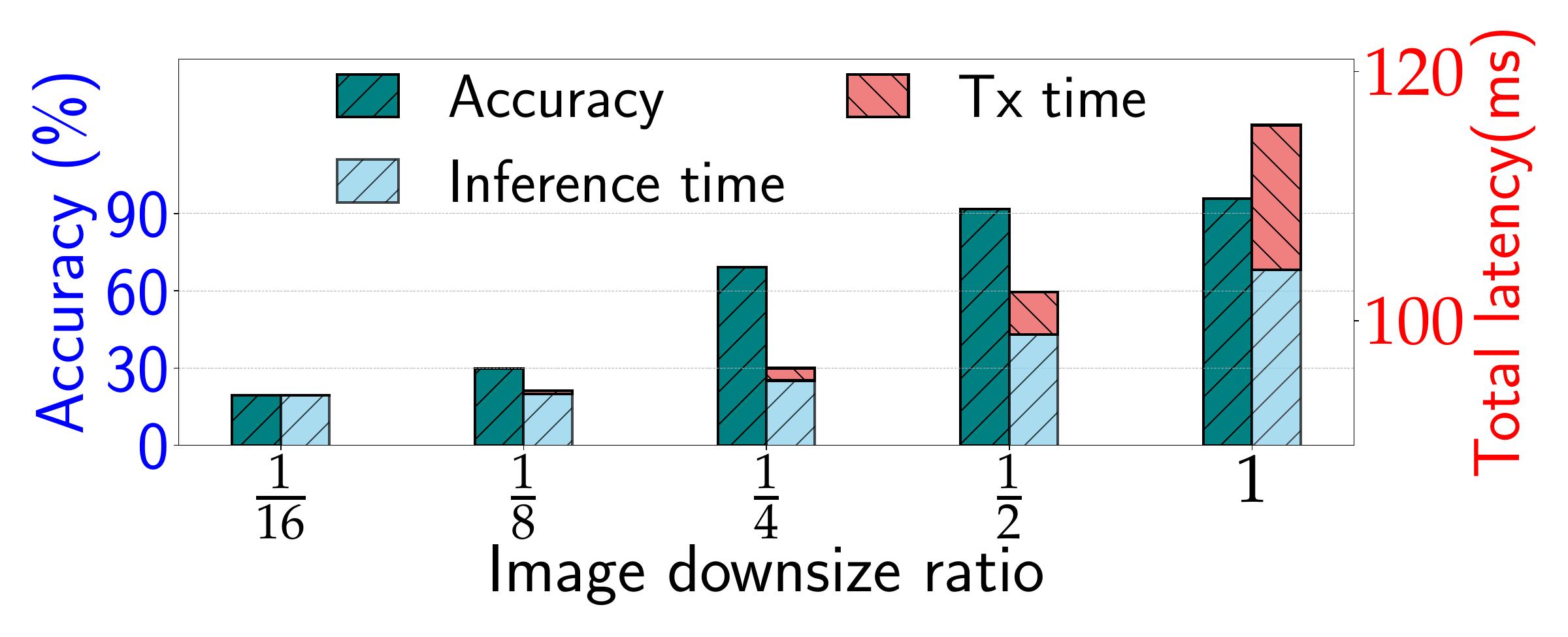}\label{fig:expression_detection_size} }
	\hfill
		\subfloat[Image compression ratio.]{%
		\includegraphics[width=\columnwidth]{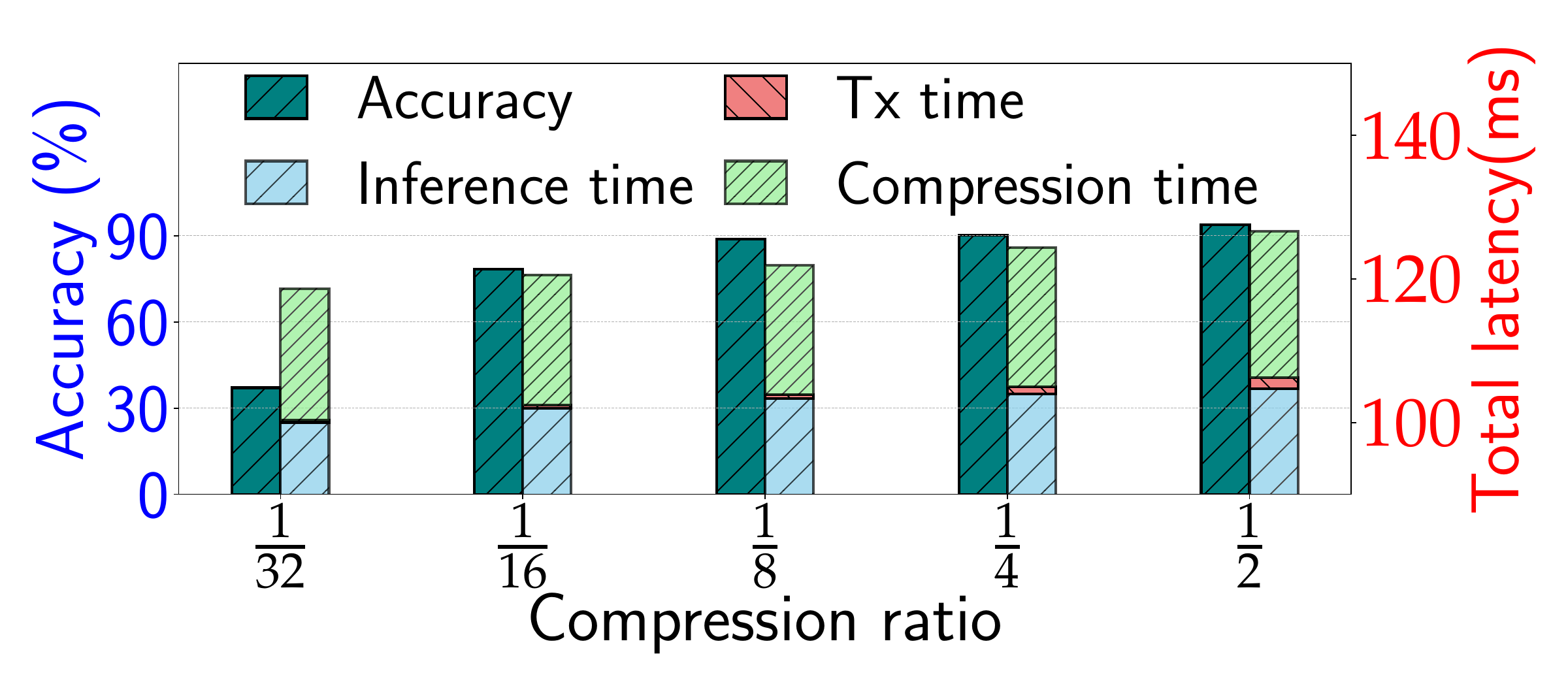}\label{fig:expression_detection_compression}}
	\setlength\abovecaptionskip{0.1cm}
 
	\caption{Performance comparison of expression detection with variations in image downsize and compression ratio. \vspace{-0.4cm}
 }
\end{figure}

Figure \ref{fig:expression_detection_size} depicts the performance of facial expression detection with various image downsize ratios i.e., various image resolutions. The results show that the \gls{dnn} accuracy for facial expression recognition degrades by up to 80\% if the images are downsized by a factor smaller than 1/2, starting from an accuracy decrease of more than 20\% if downsized by 1/4. On the other hand, transmitting the full image without applying downsampling would be infeasible due to the increase in the latency of the system. \textit{Notice that for more complex tasks such as instance segmentation, there are no available models to execute 10K resolution images directly. This is due to the unavailability of trained models and datasets for training at such a higher resolution and inherent computational complexity. To fit a 10K image, the frames need to be downsized at least by 1/16 times} which would surely fail to maintain the performance requirements. Also, note that \gls{vr} applications would need much higher resolution frames than 1024$\times$1024 thus further confirming the need for \FW~\cite{HTC-Vive-Focus-3}. The impact of compression is evaluated in Figure \ref{fig:expression_detection_compression} considering compression ratios of 1/2, 1/4, 1/8, and 1/32. For image compression, the latency should account for an additional fixed factor associated with the compression algorithm execution. Being this fixed, the total latency can only be reduced by reducing the inference time, i.e., using a more aggressive compression ratio. The results show that to achieve a latency reduction of at least 20\% a compression ratio of 1/32 should be applied. However, this leads to a reduction in the accuracy of the \gls{dnn} expression recognition algorithm by more than 60\%.

The objective of \FW is to address the challenge of maintaining high accuracy in the edge computing task execution while reducing the total latency of the system. This is achieved by integrating wireless sensing methodologies within the communication and computing system enabling the transmission of small-size but high-resolution images containing the region of interest for the algorithm execution.
Some existing work that is complementary to ours has used wireless sensing for vision-oriented approaches in multi-modal \cite{zhao2019enhancing}, cross-modal \cite{yang2022metafi}, and transfer learning settings \cite{geng2022densepose}. For example, Xie et al. \cite{xie2023mozart} utilizes a single off-the-shelf \gls{toa} depth camera to generate high-resolution maps in a noisy and dark environment. Here, a deep auto-encoder is used on top of lightweight phase manipulation functions to generate high-resolution maps. Wi-Mesh by Wang et al. \cite{wang20223d} proposes a Wi-Fi vision-based 3D human mesh development approach by estimating \gls{2d aoa} of Wi-Fi signals. 3D mesh is obtained from two \gls{2d aoa} estimated from two different receivers placed at two different locations. This approach integrates the location information of the camera with multiple \gls{ap} readings. It assigns weights to different \gls{ap} readings by analyzing the user's orientation information obtained from the camera. \textit{However, to the best of our knowledge, none of the earlier work has proposed wireless sensing to assist edge offloading.} \vspace{-0.1cm}

\begin{figure}[t]
    \centering
    \includegraphics[width=\columnwidth ]{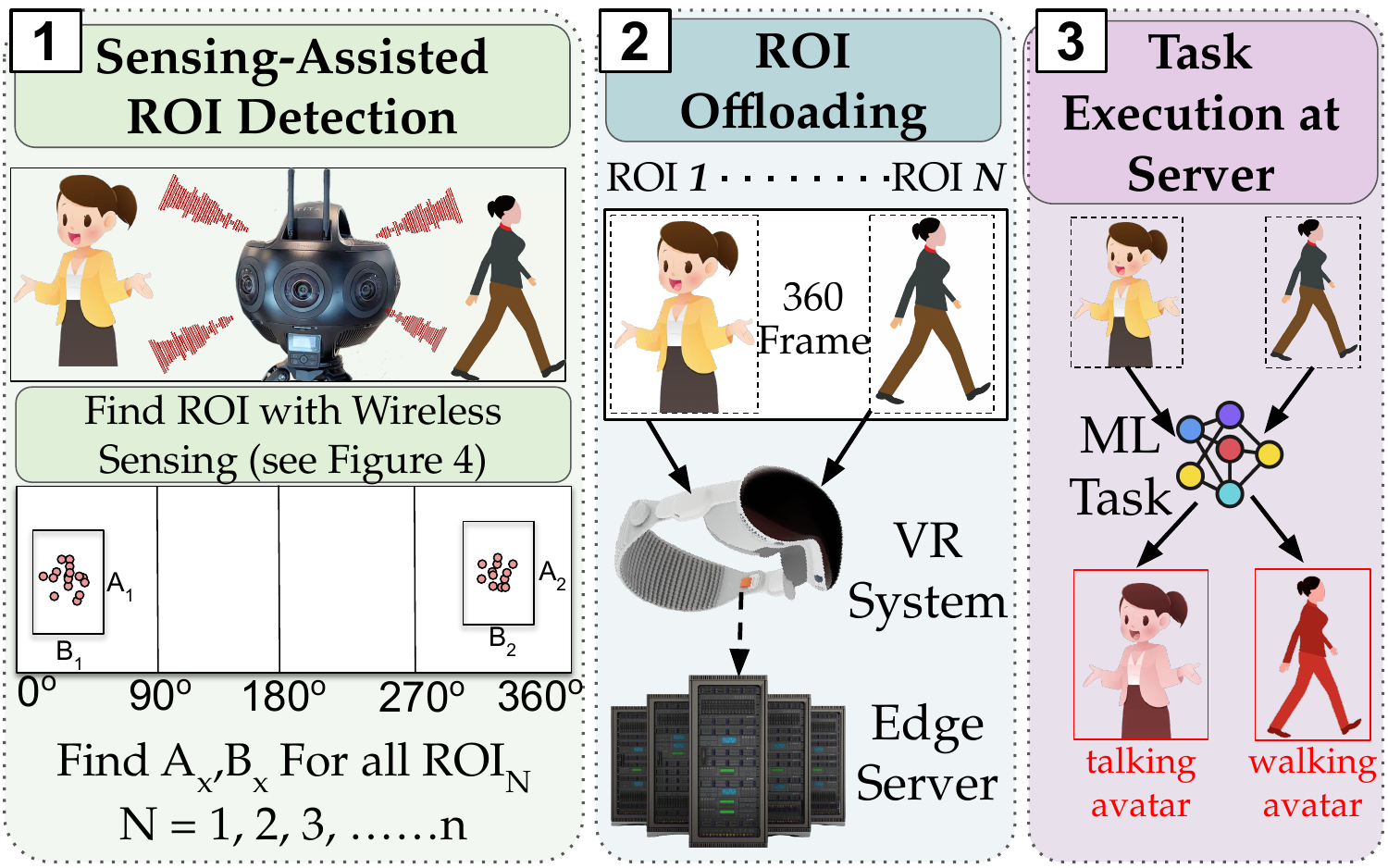}
    \setlength\abovecaptionskip{-0.1cm}
    \caption{Overview of the \FW Framework. \vspace{-0.3cm}}
    \label{fig:SAWEC_overview}
\end{figure}

\section{System Overview} \label{sec:system_overview}

\FW empowers modern mobile devices featuring \cam cameras with wireless sensing functionalities that identify the \gls{roi} inside each captured video frame before transmitting it to the edge. In this view, the wireless \gls{nic} at the camera device serves both for image data offloading to the edge server and for wireless sensing parameter estimates for \FW image processing. Specifically, \FW consists of three main blocks, as depicted in Figure~\ref{fig:SAWEC_overview} and summarized below: (1) sensing-assisted \gls{roi} detection, (2) \gls{roi} offloading, and (3) task execution at the edge server.

\smallskip
\textbf{(1) Sensing-assisted ROI Detection:} The detection of the \gls{roi} relevant for the edge computing task execution is based on the context information obtained through wireless sensing. Specifically, we leverage the \gls{cfr} estimated by the wireless \gls{nic} to detect the dynamics in the environment. \FW synchronizes channel measurements with the video frames through their timestamps, and obtains an estimate of the locations of the targets in the environment by processing the \gls{cfr} through multi-path parameter estimation algorithms. The combination of clustering and tracking algorithms allows selecting the \glspl{roi}, being the areas where changes were detected. Note that the \gls{cfr} estimation is performed by leveraging the training field in received data packets. Therefore, the system requires at least another device in the network. Without loss of generality, this can be the \gls{ap} that provides connectivity to the camera. Hereafter we will consider that two (the minimum) devices are involved and will refer to the \textit{sensing \gls{nic}} as the \gls{nic} where the \gls{cfr} estimation is performed and to the \textit{transmitter \gls{nic}} as the one transmitting wireless signals for triggering \gls{cfr} estimation.\vspace{0.1cm}


\smallskip
\textbf{(2) ROI Offloading:} The \gls{roi} offloading block is in charge of selecting the portion of the high-resolution image frame to be offloaded to the edge server based on the context information gained through wireless sensing. By transmitting only the \glspl{roi} instead of the entire frame, \FW reduces the wireless channel occupation and the transmission latency while guaranteeing high performance for the computing task execution at the edge.

\smallskip
\textbf{(3) Task Execution at Edge Server:}~The edge server receives the \glspl{roi} from the camera and uses them as input for the \gls{dnn} task, thus obtaining two main advantages with respect to traditional approaches. First, by executing the \gls{dnn} on high-resolution \glspl{roi} instead of on their compressed and/or reshaped versions, \FW reaches higher accuracy than other edge computing task offloading methods in the literature. Moreover, processing the \glspl{roi} instead of the whole (bigger) frames reduces both the training and inference time. Combined with the reduced transmission latency mentioned above, \FW allows reducing the end-to-end transmission and computing time. Notice that, if the detected \gls{roi} is too large to directly fit into the \gls{dnn} architecture, \FW reshapes the \gls{roi} to the maximum allowed size. However, by using our proposed wireless sensing algorithm the reshaping ratio is about one, indicating minimal information loss for the reshaping process. \vspace{-0.1cm}



\section{Sensing-Aided ROI Detection}\label{section: Wi-Fi-aided ROI Detection}

The \gls{cfr}-based localization has been implemented by adapting the super-resolution multi-path parameter estimation algorithm MD-Track~\cite{Xie2019MDTrack}. For \glspl{roi} detection and tracking, we implemented a custom-tailored approach based on density-based spatial clustering of applications with noise (DBSCAN). 

\begin{figure}[t]
    \centering
    \includegraphics[width=\columnwidth ]{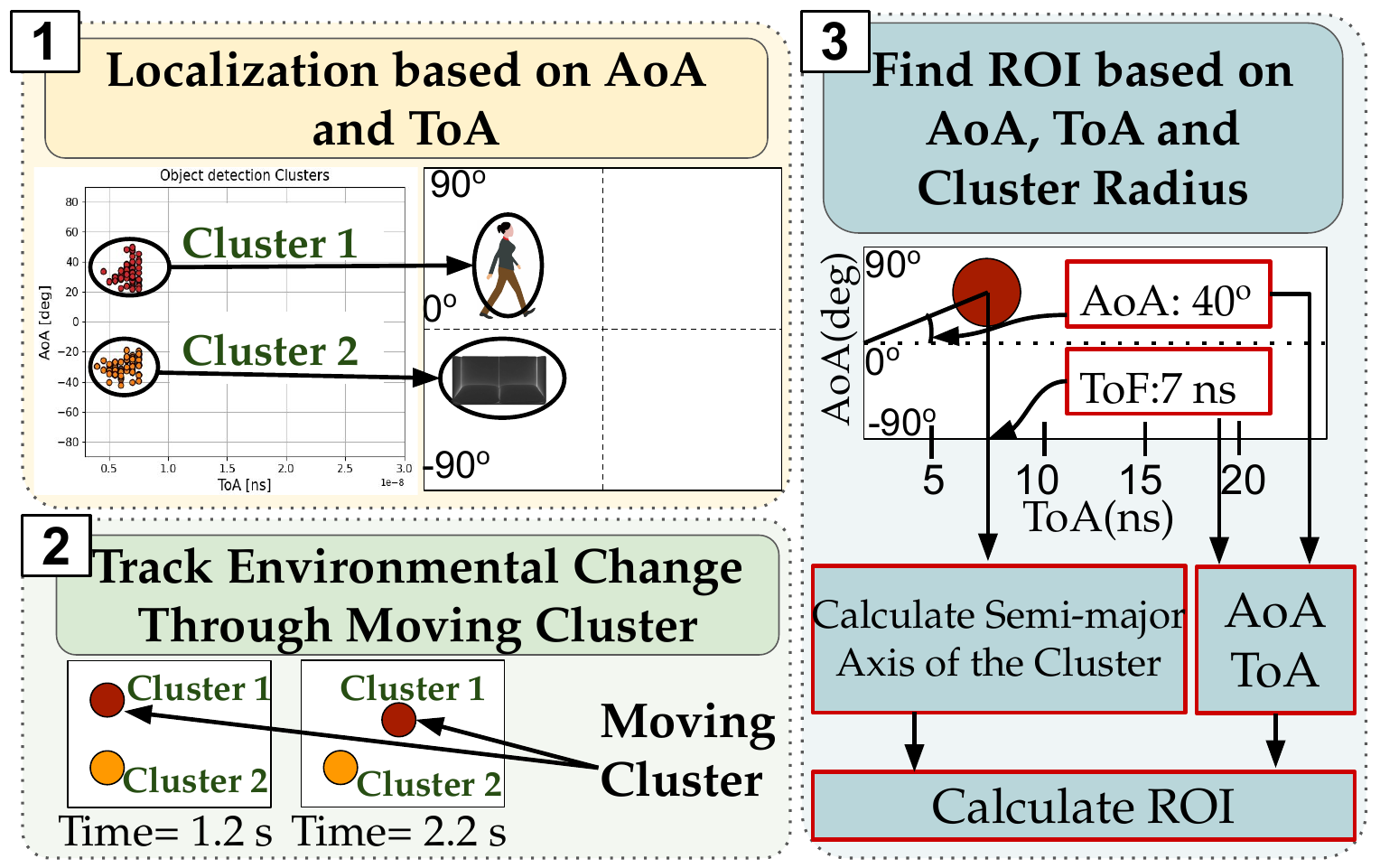}
    	\setlength\abovecaptionskip{-0.05cm}
    \caption{Detection of the Region of Interest (ROI). \vspace{-0.1cm}}
    \label{fig:ROI_detection}
\end{figure}

\subsection{Wi-Fi Channel Model}

We consider a $1 \times N$ IEEE 802.11ax (Wi-Fi) multi-antenna system where $N$ is the number of receiving antennas and $n \!\in\! \{0, \dots, N \!-\! 1\}$ indicates the receiving antenna index. Being $f_c$ the main carrier frequency, $\Delta f$ the \gls{ofdm} sub-channel spacing and $T\!=\!1/\Delta f$ the \gls{ofdm} symbol time, the \gls{cfr} for sub-channel \mbox{$k \in \{-K/2, \dots, K/2 - 1\}$}, estimated at receiver antenna $n$ and time $t$, $H_{k, n}(t)$, is  modeled as~\cite{goldsmith_2005}
\beq \label{eq:h}
	 H_{k, n}(t) = \sum_{p=0}^{P-1} A_{p}(t) e^{-j2\pi (f_c + k/T)\tau_{p, n}(t)}\;,
\eeq
where $p \in \{0, \dots, P - 1\}$ represents the $P$ multi-path components associated with the wireless signal propagation, each of which is characterized by an attenuation $A_{p}(t)$ and a \gls{toa} $\tau_{p, n}(t)$ (also referred to as propagation delay).
Each multi-path component $p$ is associated with a static or moving object in the environment that acts as a reflector, diffractor, or scatterer for the wireless signal propagating from the transmitter to the receiver. The propagation delay $\tau_{p, n}(t)$ is associated with the position $\ell_{p}(t)$ of the $p$-th object in the environment and the collecting antenna. Each multi-path component is collected by each antenna in subsequent time instants that depend on the \gls{aoa} $\theta_{{\rm rx},p}(t)$. Indicating with $\Delta_{p, n}^{\rm rx}(t)$ the antenna-dependent contribution to the length of the $p$-th component, the \gls{toa} is obtained as 
\beq \label{eq:tau_p}
    \tau_{p, n}(t) = \frac{\ell_{p}(t)  + \Delta_{p, n}^{\rm rx}(t)}{c}\;,
\eeq 
where $c$ is the speed of light. Considering a linear array with antennas spaced apart by $d_{\rm rx}$ and using the left antenna as the reference, we have $\Delta_{p, n}^{\rm rx}(t) = n \sin (\theta_{{\rm rx},p}(t)) d_{\rm rx}$, where the \gls{aoa} $\theta_{{\rm rx},p}(t)$ is measured clockwise starting from the direction perpendicular to the antenna array.

\subsection{Super Resolution Multi-Path Parameter Estimation}\label{subsection:super_resolution}

To ease the proper estimation of the \gls{aoa}, the receiver antennas are usually spaced apart by $d_{\rm rx}=\lambda/2$ where $\lambda= c/f_c$ is the wavelength of the carrier signal. In turn, $2(f_c + k/T) d_{\rm rx}/c \simeq 1$ and the contribution of $\Delta_{p, n}^{\rm rx}(t)$ to $H_{k, n}(t)[p]$ can be written as $e^{-j\pi n \sin (\theta_{{\rm rx},p}(t))}$. 
The parameters of the multi-path components in $\tau_{p, n}(t)$, i.e., $\ell_{p}(t)$, $\theta_{{\rm rx},p}(t)$, can be estimated by computing the inverse Fourier transformation of the \gls{cfr} in \eq{eq:h} over the subcarrier $k$ and receiver antenna $n$ dimensions, respectively. However, the resolution of the Fourier transform approach is constrained by the \gls{cfr} diversity in the frequency and spatial domains. Super-resolution algorithms can be used to deal with these limitations and obtain more precise \gls{aoa} and \gls{toa} estimates by leveraging the sparse nature of the wireless channel~\cite{meneghello2023wi}. 
The intuition behind super-resolution algorithms is that in case two paths cannot be separated considering one of the parameters, they may be separable in the other dimension. For example, in the case of two paths closely spaced in time, they may be characterized by different \glspl{aoa} and, in turn, can be identified as two separate contributions. Hence, the resolution can be improved by jointly estimating all the parameters of each multi-path component. In this work, we use the iterative mD-Track algorithm for this purpose~\cite{Xie2019MDTrack}. Note that we focus on the identification of the \gls{aoa} and \gls{toa} ($2$-dimensional mD-Track) as they are sufficient for providing enough side information for \gls{roi} detection.\vspace{-0.2cm}


\subsection{\gls{aoa} and \gls{toa} Clustering and Tracking}\label{subsec:clustering}
\vspace{-0.1cm}
To improve the accuracy of the local pre-processing (frame windowing) before frame transmission, \FW uses a number of channel estimated for each frame. Specifically, being $V$ the video frame collection rate, measured in \gls{fps}, and $C$ the frequency (in Hz) of \gls{cfr} collection (and, in turn \gls{aoa} and \gls{toa} estimation), \FW set $C \ge V$ such that the localization for each frame is performed using $C/V \ge 1$ \gls{cfr} samples.
For example, considering a system with \gls{cfr} estimation frequency of 200 Hz, and video frame rate of 25 \gls{fps}, the \gls{aoa} and \gls{toa} estimation for each frame is performed leveraging $200/25 = 8$ \gls{cfr} samples.
The localization proceeds as summarized in Figure~\ref{fig:ROI_detection}. At first, the \gls{aoa} and \gls{toa} are estimated for each of the $C/V$ \gls{cfr} samples related to a video frame. After that, the $C/V$ \gls{aoa}/\gls{toa} pairs are clustered through \gls{dbscan}, removing the outliers associated with noise (step \textbf{1}). Hence, the centroids are compared with the centroids of the clusters associated with the previous video frame to detect any change in their location (step \textbf{2}). The information about moving targets is then used in step \textbf{3} to obtain the \gls{roi} for video frame processing, as detailed in Sections~\ref{sec: adaptation} and \ref{subse:roidetect}.\vspace{-0.1cm}


\vspace{-0.2cm}
\subsection{\gls{aoa} and \gls{toa} Projection into the Camera Reference System}
\label{sec: adaptation}
\vspace{-0.1cm}


\begin{figure}[t]
	\centering
	\subfloat [\gls{nic} on the right ($A_x > B_x$).]{%
		\includegraphics[width=.48\columnwidth]{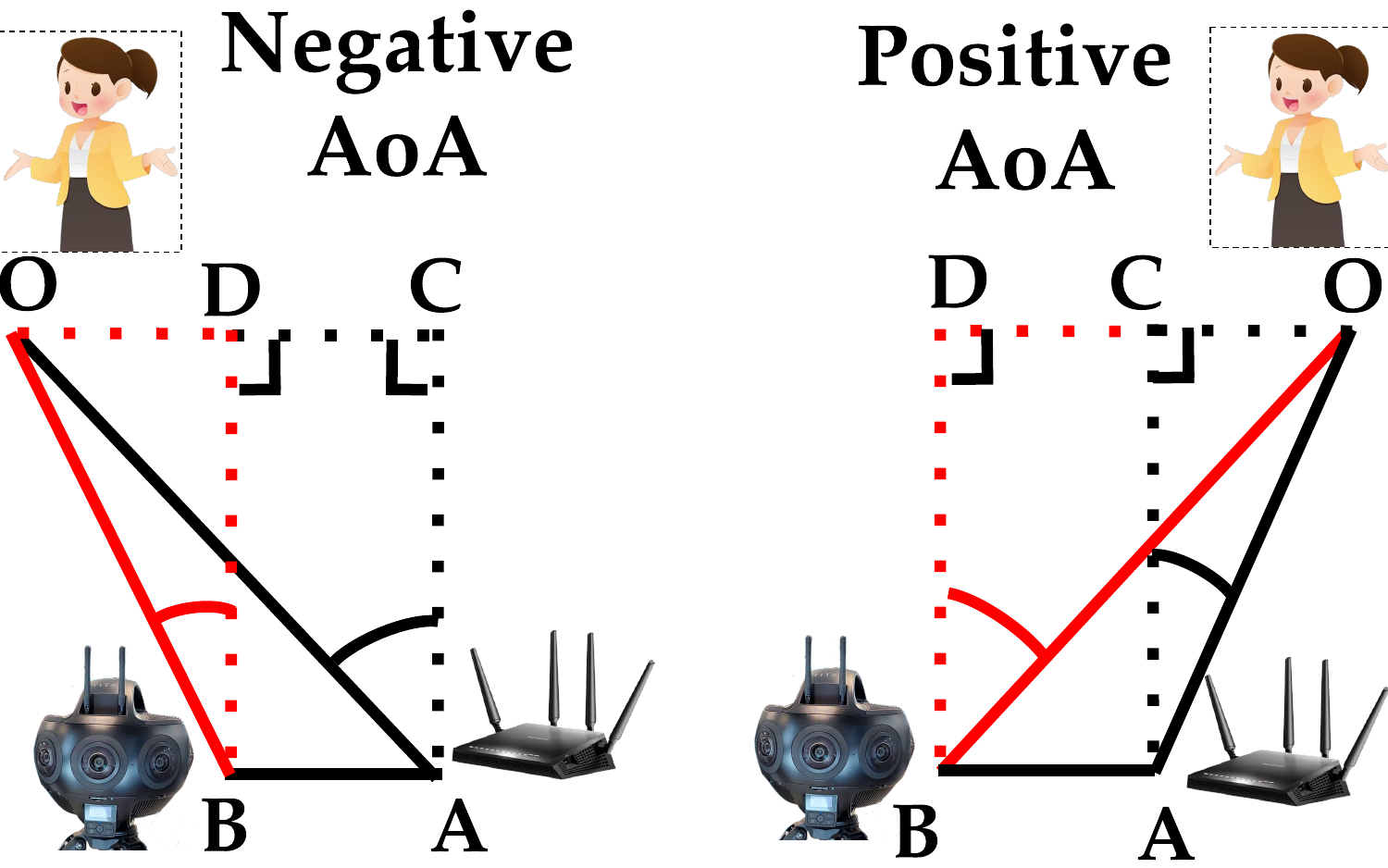}}
	\hfill
		\subfloat[\gls{nic} on the left ($A_x < B_x$).]{%
		\includegraphics[width=.48\columnwidth]{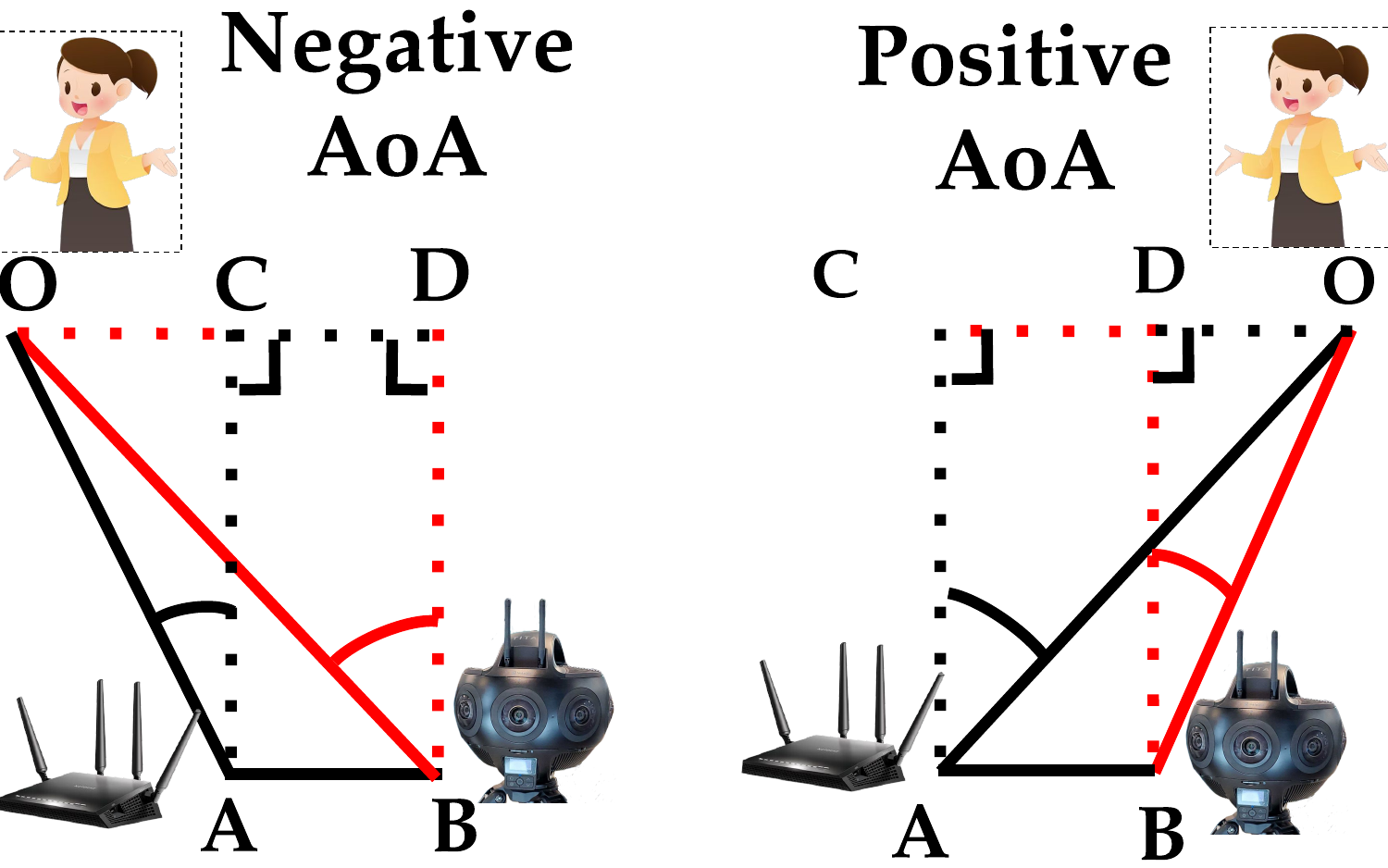}}
	\setlength\abovecaptionskip{0.1cm}
 
	\caption{\gls{aoa} and \gls{toa} projection into the camera reference system. $AB$ is the distance between the \mbox{Wi-Fi} NIC and the camera. $\angle OAC$ and $OA$ are respectively the \gls{aoa} and range estimated through Wi-Fi sensing. $\angle OBD$ and $OB$ are the \gls{aoa} and range projected into the camera reference system.
 \vspace{0.2cm} 
 }
	\label{fig: aoa_adaptation} 
\end{figure}

For proper selection of the \gls{roi} through the sensing side information, the  \gls{nic} for \gls{cfr} data collection and the \cam camera have to share the same reference system. This would require: (i) the camera and the sensing \gls{nic} to be exactly co-located and, in turn, capture the same field of view, and (ii) the \gls{nic} transmitting opportunistic signals to trigger \gls{cfr} estimation to be placed exactly on the tangent of the sensing \gls{nic} antenna array, to share the same common zero of \gls{aoa} reference frame.
However, these requirements are hardly achievable from a physical perspective. In turn, we need to `project' the \gls{aoa} and \gls{toa} estimated through the  sensing \gls{nic} into the camera reference system. For this, we designed a procedure consisting of the two phases detailed next. \smallskip

\textbf{Phase 1} At first, we apply a correction for the slightly different placement of the sensing \gls{nic} and the camera. We reasonably assume that they are aligned on two of the three 3D axes (Y and Z axes), and we indicate with \(A_x\) and \(B_x\) their respective X-axis location. In Figure~\ref{fig: aoa_adaptation} we provide an overview of the different respective positions of the \mbox{Wi-Fi} sensing \gls{nic}, the camera, and the target. Based on these possible situations we designed a corrective algorithm to obtain $\angle OBD$ (\gls{aoa} in the camera reference system) from $\angle OAC$ (\gls{aoa} in the sensing \gls{nic} reference system). This is obtained by applying trigonometrical transformations as detailed in Algorithm~\ref{alg: adaptation}. Note that \gls{aoa} $\angle OAC= 0$ means that the object is on the tangent connecting the object and the \gls{nic} antenna array ($AC$ in Figure~\ref{fig: aoa_adaptation}). \Gls{aoa} $\angle OAC\ge 0$ means that the object is on the right with respect to the tangent, while \mbox{\gls{aoa} $\angle OAC\le 0$} is associated with an object on the left.\smallskip

\begin{figure*}[t]
    \centering
    \includegraphics[width=0.95\textwidth]{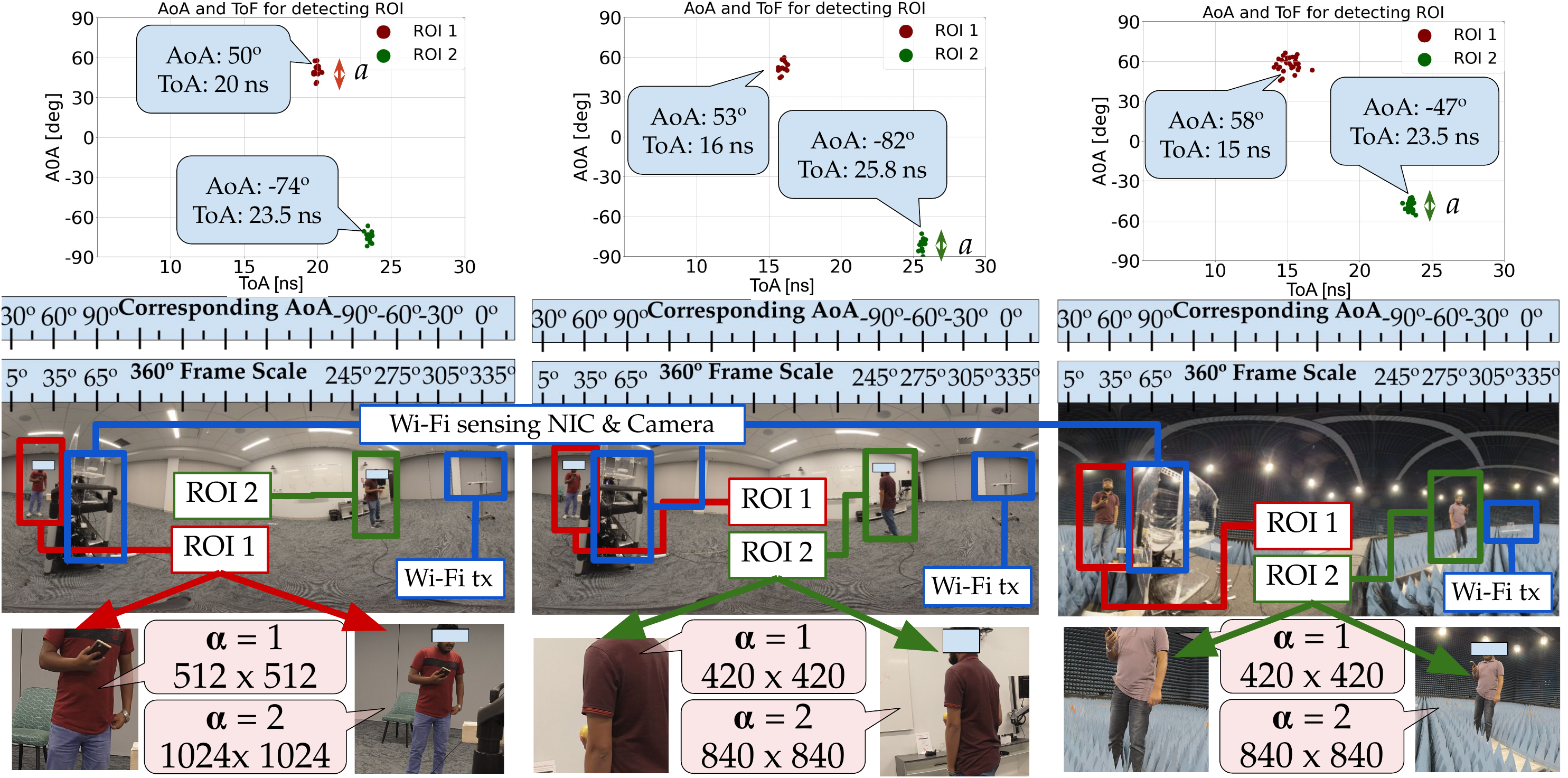}
    \caption{\Gls{roi} detection from \gls{aoa}/\gls{toa} estimation and clustering. The \gls{roi} multiplying factor $\alpha$ is the parameter used for obtaining the size of the \gls{roi} from the extension of the cluster ($a$) to account for localization errors.
    \vspace{-0.1cm}}
    
    
    \label{fig:frame_translation}
\end{figure*}

\textbf{Phase 2} To address the challenge related to the misplacement of the transmitter \gls{nic} with respect to the tangent and obtain the same reference for the zero \gls{aoa} frame, we first set a \gls{aoa} scale for the \cam ~frame assigning 0$^{\circ}$ to the left edge of the camera field of view. 
Hence, we project the \gls{aoa} $\angle OBD$, obtained above, into the \gls{aoa} value in the new reference system, hereafter referred to as $\theta$. Specifically, we have $\theta = [(\angle OBD + \theta_{\rm tx}) ~~{\rm mod} ~~360]$ where ${\rm mod}$ represents the modulo operation and $\theta_{\rm tx}$ indicates the location of \mbox{Wi-Fi} transmitter in the 360$^{\circ}$ reference system. An example of the processing is depicted in Figure \ref{fig:frame_translation} using real experimental data. For example, $\angle OBD = 50^\circ$ identifying \gls{roi} 1 corresponds to $\theta = 25^\circ$.\vspace{-0.15cm}

\begin{algorithm}[t]
\caption{\gls{aoa} and \gls{toa} projection into the camera reference system. The geometrical model is in Figure \ref{fig: aoa_adaptation}.}\label{alg: adaptation}
\begin{algorithmic}
\Require $A_x, B_x, \angle OAC, OA$ 
\Ensure $\angle OBD, OB$
    \State \(OC = OA \cdot \sin(\angle OAC)\) 
    \State $CD = AB$ 
    \State \(BD =  AC =  OA \cdot \cos(\angle OAC)\)  
    \State $\zeta_{\rm pos} \gets {\rm sgn}(A_x - B_x)$ \Comment{$ \zeta_{\rm pos}\!=\!1$ if camera is on the left of NIC}
    \State $\zeta_{\rm aoa} \gets {\rm sgn}(\angle OAC)$ \Comment{$ \zeta_{\rm aoa}\!=\!1$ if positive \gls{aoa}}
    \State $OD = OC + \zeta_{\rm pos} * \zeta_{\rm aoa} * CD$
    \State $\angle OBD = \arctan \left(\frac{OD}{BD}\right)$; 
    $OB = \left(\frac{OD}{\sin\angle OBD}\right)$
\end{algorithmic}
\end{algorithm}
\setlength{\textfloatsep}{0.2cm}

\subsection{Estimating the Size of the \gls{roi}}\label{subse:roidetect}

For each video frame, we leverage the extension of the detected clusters (see Section~\ref{subsec:clustering}) in the \gls{aoa} domain to derive the size for the \glspl{roi}. Hereafter we will refer to such extension, i.e., the ranges of \gls{aoa} covered by the cluster, as $a$ (measured in degrees). The center of each \gls{roi} is set to the center of the cluster and to define its extension we introduce a parameter, hereafter referred to as \gls{roi} multiplying factor and indicated with $\alpha$, to account for localization error. Specifically, the \gls{roi} extension is set to $a \cdot \alpha^\circ$ in the \gls{aoa} domain. For ease of processing, we considered the same extension in the width and height of the \gls{roi}. This means that we obtain the number of pixels to be considered for \gls{roi} height \& width by using $a\cdot\alpha\cdot W/360$ where $W$ is the width of the entire frame.

As shown in the bottom part of Figure \ref{fig:frame_translation}, for our specific setup, $\alpha=1$ is too small for having the whole subject in the \gls{roi} whereas $\alpha=2$ allows retaining the entire subject.\vspace{-0.1cm}

\subsection{\FW Processing Time}
\begin{figure}[h]
	\centering
		\includegraphics[width=\columnwidth]{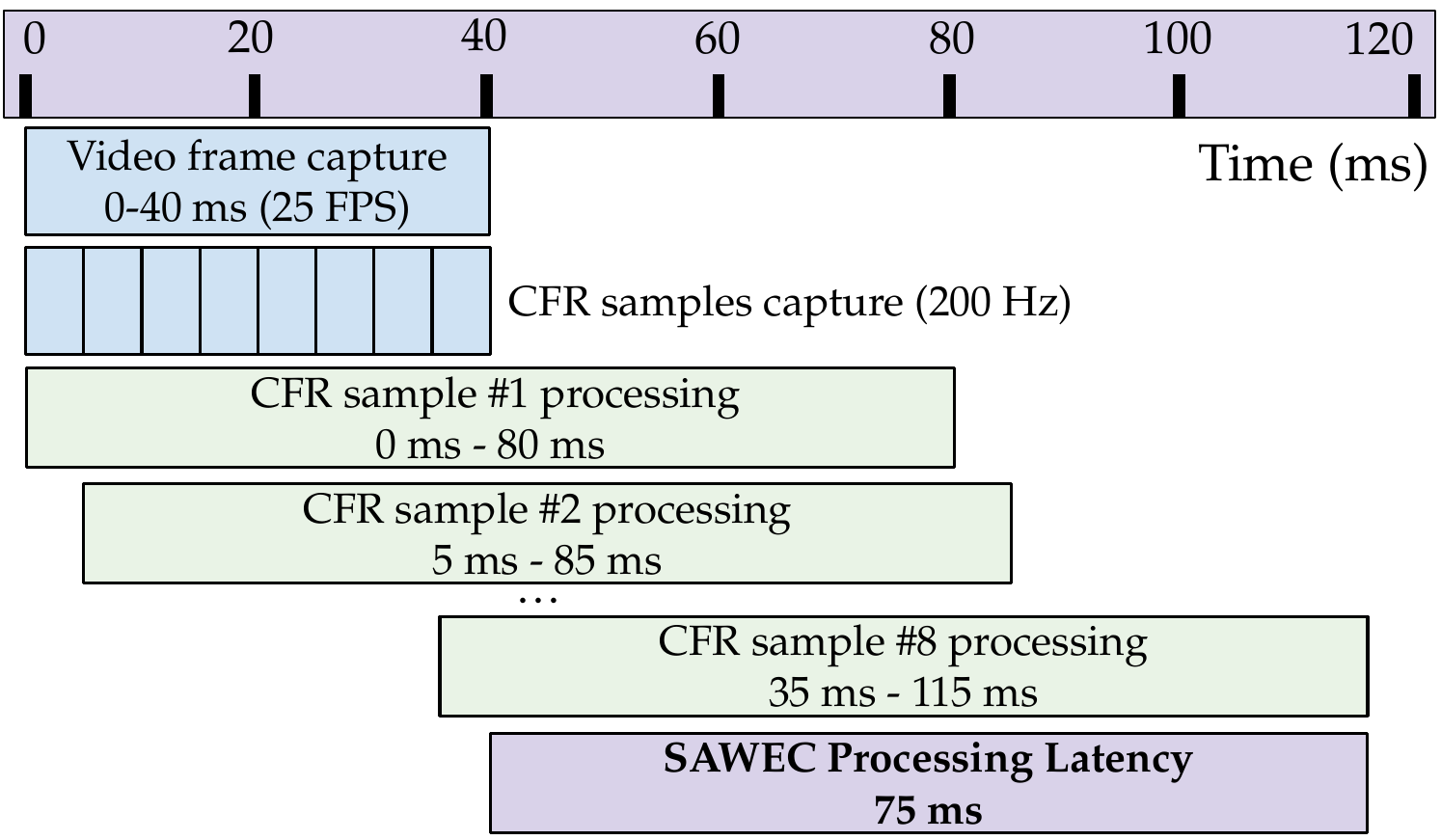}
  \setlength\abovecaptionskip{-0.05cm}
	\caption{End-to-end Processing Latency of \FW.}

 
	\label{fig:SAWEC_processing_time} 
\end{figure}

\FW considers $C/V$ \gls{cfr} samples (with $C$~Hz sampling frequency) for estimating \gls{aoa} and \gls{toa} for each video frame (at $V$~FPS). 
Indicating with $T$ the processing time for a \gls{cfr} sample, the total \FW processing time is defined by $T + (C/V -1) \times 1/C$ and \FW processing latency is defined by $(T + (C/V -1) \times 1/C) - 1/V$. For example, considering a system with  $C = 200$~Hz, $V = 25$~FPS and $T = 80$~ms, \FW overall processing time is $115$~ms and latency is $75$~ms, as depicted in Figure~\ref{fig:SAWEC_processing_time}.
Here, one video frame at 25 FPS corresponds to 8 \gls{cfr} samples at \gls{cfr} estimation frequency of 200 Hz. The first packet is received at 0 ms and the next ones are at an interval of 5 ms. A single \gls{cfr} sample requires 80 ms for processing, while the combined processing time for all 8 \gls{cfr} samples associated with a video frame totals 115 ms. Consequently, the \FW processing latency amounts to 75 ms as the processing time for a single video frame is 40 ms at 25 FPS.\vspace{-0.2cm}

\section{Experimental Setup}\label{sec:perf-ev}
We evaluated \FW by implementing the system on commercial devices available on the market. The system comprises an Insta360 Titan 360$^\circ$ camera for video capturing, a Linux machine for video processing, and an IEEE 802.11ax network for communication and wireless sensing (localization and tracking). The system was configured to capture frames at a resolution of 10K with a frame rate of 25 \gls{fps}. The IEEE 802.11ax network comprised two \gls{cots} AX200 \gls{nic} operating on a 5~GHz Wi-Fi channel with 160MHz of bandwidth. One and two antennas were enabled at the transmitter and sensing devices respectively. At the sensing Wi-Fi \gls{nic}, we used PicoScenes for \gls{cfr} collection~\cite{jiang2021eliminating}. We placed the Wi-Fi receiver and the 360$^\circ$ camera in closed proximity and synchronized the \mbox{Wi-Fi} localization and camera systems by using their internal reference clock. The Wi-Fi transmitter was placed on the opposite side to properly irradiate the environment and obtain valuable information for sensing. We considered two different propagation environments for the evaluation: (i) an entrance hall and (ii) an anechoic chamber. While the entrance hall allows evaluating the performance of \FW in a real-world environment, the anechoic chamber provides the best-case scenario evaluation as the multi-path effect is reduced. 
Note that although our experimental evaluation considers the \mbox{Wi-Fi 6} (IEEE 802.11ax) standard, the proposed system and the related analysis are general and can be applied to wireless standards operating with \gls{mimo}.

\begin{figure}[t]
    \centering
    \includegraphics[width=0.92\columnwidth]{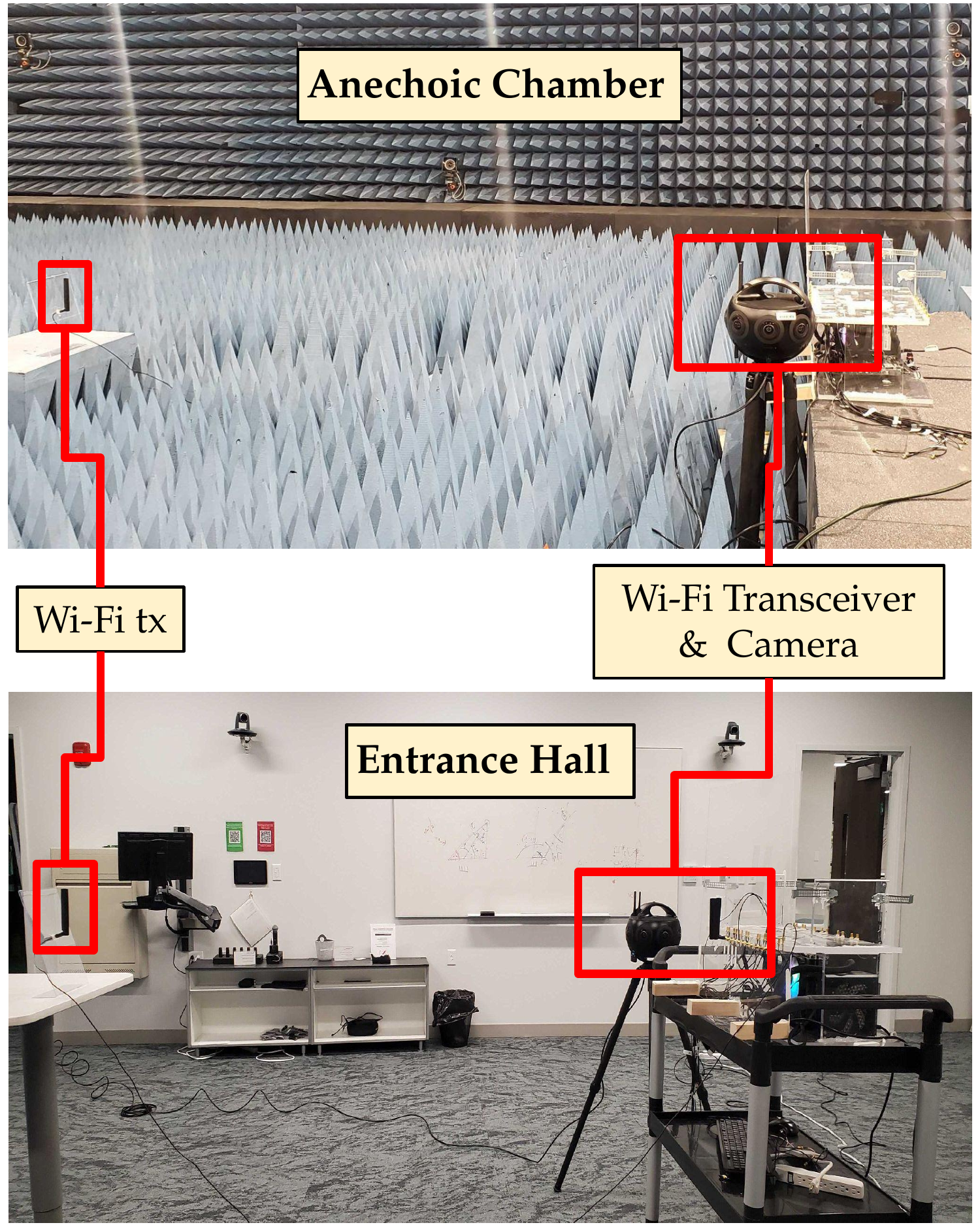}

    \includegraphics[width=0.92\columnwidth]{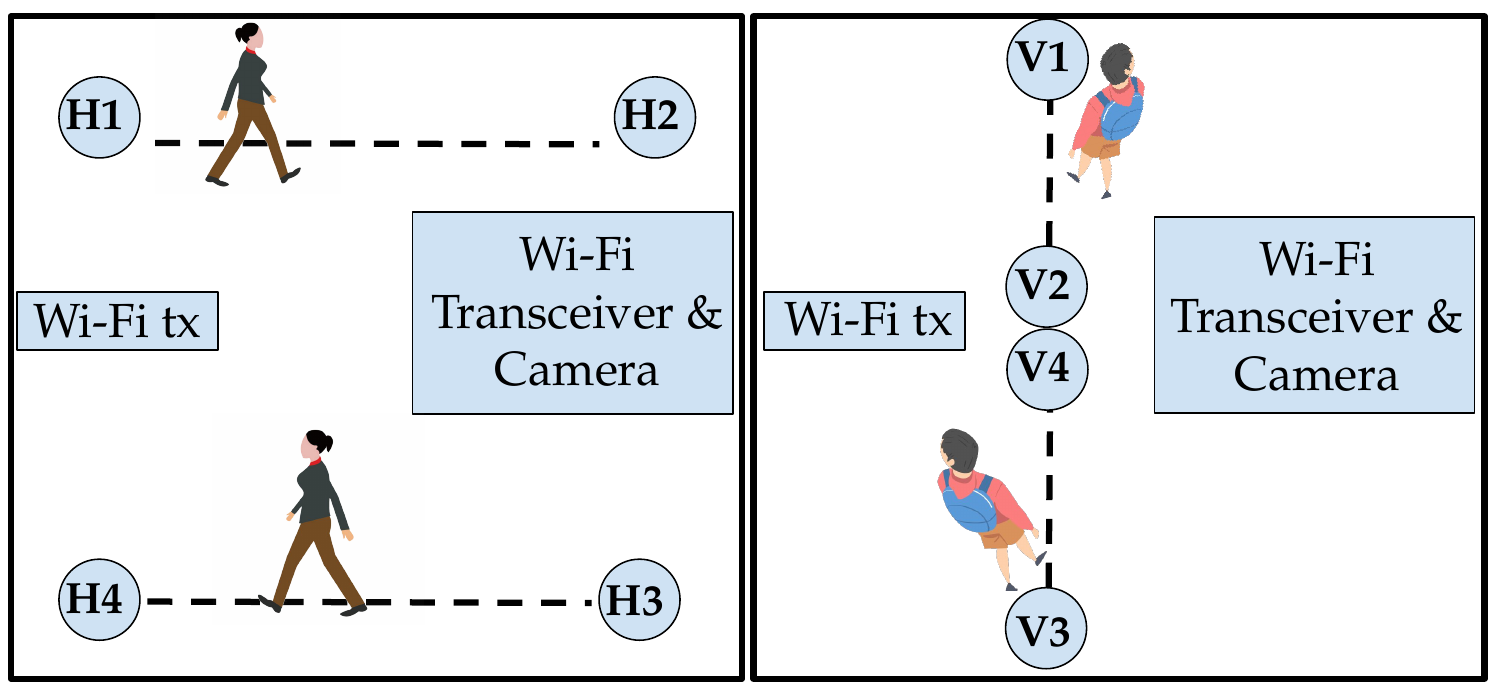}
    \caption{Experimental setup and \FW evaluation scenarios.\vspace{-0.1cm}
    }
    \label{fig:Experimental_setups}
\end{figure}

The experimental setup along with the evaluation scenarios are presented in Figure \ref{fig:Experimental_setups}. We performed six different tests in each of the environments as depicted in Figure \ref{fig:Experimental_setups}: a single person walking between (i) H1 to H2 (ii) H3 and H4 (iii) V1 and V2 (iv) V3 and V4, and two persons walking simultaneously between (v) H1 and H2 and H3 and H4 and (vi) V1 and V2 and V2 and V3. Data were collected for three minutes for each of the tests. The tests on the anechoic chamber and the entrance hall were performed on different days, at different times of the day and with different orientations of the systems. Due to physical constraints, the 360$^\circ$ camera is negatively shifted along the x-axis by 1~m in the hall room and 0.5~m in the anechoic chamber whereas the spacing between the Wi-Fi transceiver and the \gls{cfr} injector is 3~m for both the environments. 

\vspace{2cm}
\section{Performance Evaluation}\label{sec:exp-res}
We assessed the performance of \FW by performing an extensive experimental data collection campaign and evaluating three main metrics: (i) accuracy of the \gls{dnn} task; (ii) wireless channel occupation; and (iii) end-to-end latency. In the following results, the metrics for the anechoic chamber and the entrance hall have been averaged over all the frames captured in the two evaluation environments. For a baseline comparison, we evaluate \FW against two \gls{sota} algorithms: (i) YolactACOS adaptive edge assisted segmentation \cite{xie2022edge}, (ii) EdgeDuet context-aware data partition-based \gls{wec} \cite{yang2022edgeduet}.

\textbf{YolactACOS} approach is based on a selective offload of the computing tasks. Lightweight tasks are processed directly on the mobile devices to ensure responsiveness and minimize latency while more computationally intensive tasks are offloaded to nearby edge servers. This workload subdivision allows the system to execute near real-time computer vision tasks without overburdening the limited computational resources at the mobile devices. However, the execution of deep learning models (even if they are lightweight) on mobile devices and the need to offload high-resolution full frames to the edge servers for computationally intensive tasks negatively affecting mobile device energy consumption, end-to-end latency, and overall task performance as shown in Section \ref{sec:exp-res}.

\textbf{EdgeDuet} employs a partition-based strategy to enhance the detection of small objects by distributing computing tasks between mobile devices and edge servers. This method utilizes image tiling, which entails dividing the frame into several segments, or tiles. Such tiles are then simultaneously offloaded to multiple edge servers, thus accelerating the detection process. However, this strategy requires mobile devices to run deep learning models and perform image pre-processing before offloading the tasks to the edge servers, which increases the mobile device's energy consumption and latency. Additionally, compressing the image tiles prior to offloading and the chance that a single object appears across multiple tiles further hampers the task performance, as shown in Section \ref{sec:exp-res}.

\begin{figure}[t]
	\centering
\includegraphics[width=\columnwidth]{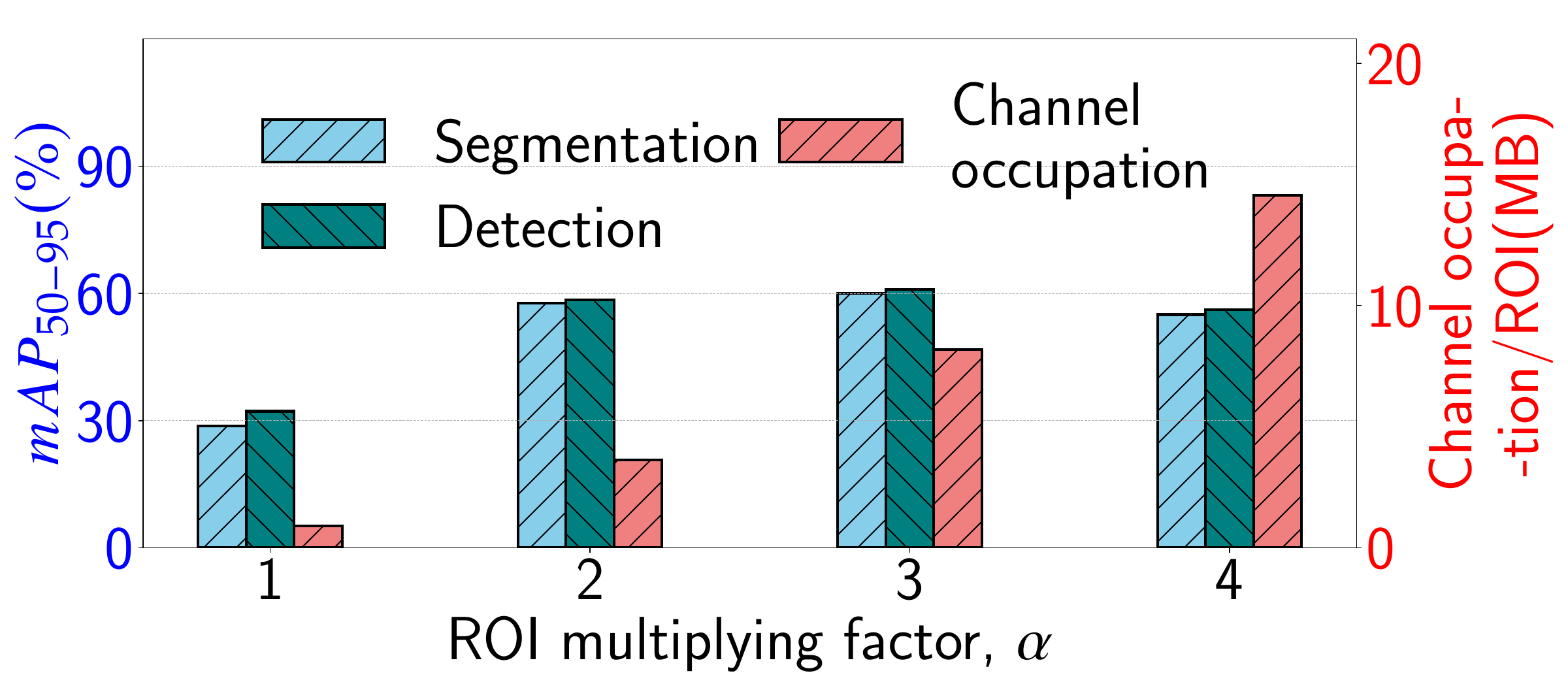}
	\setlength\abovecaptionskip{-0.2cm}
	\caption{$mAP_{50-95}$ and channel occupation per \gls{roi} for the image segmentation and object detection tasks performed on \glspl{roi} with different \gls{roi} multiplying factors ($\alpha$).\vspace{-0.2cm}
 }
	\label{fig:segmentation_performance_alpha} 
\end{figure}

In our evaluation, we considered \textit{instance segmentation} and \textit{object detection} as the computing tasks as they are widely used benchmarking tasks for edge computing. As for the \gls{dnn}, we utilized different variants of YOLOv8~\cite{Jocher_YOLO_by_Ultralytics_2023} for benchmarking our tasks. YOLOv8 represents the most recent update in the YOLO~\cite{redmon2016you} series of instance segmentation and object detection models. While retaining the foundational architecture of its predecessor, YOLOV8 incorporates several advancements. Notably, it features a novel neural network design that integrates both the feature pyramid network (FPN) and path aggregation network (PAN). FPN methodically diminishes the spatial resolution of the input image while augmenting the feature channels. This process generates feature maps that adeptly identify objects across various scales and resolutions. Conversely, the PAN strategy enhances the model's ability to discern multi-scale and resolution features by amalgamating features across different network levels via skip connections. Such capabilities are essential for the precise detection of objects of different sizes and shapes.

We implemented the edge computing server -- where the \gls{dnn} training and inference are executed -- through a Linux machine with 12th Generation Intel(R) Core(TM) i7-12700K \gls{cpu} with 64 GB of \gls{ram} (memory speed: 4800 mega-transfer/second) and a \gls{gpu} (RTX A4000) of 16 GB memory with computational capability of 7.5. We considered the $mAP_{50-95}$ performance metric to assess instance segmentation performance. This metric represents the mean average precision across \gls{iou} thresholds spanning from 50\% to 95\% where \gls{iou} measures the overlap between predicted and ground truth regions. The communication between the Wi-Fi-aided camera system and the edge computer is obtained through an IEEE 802.11ax $4\times2$ MIMO link.

\begin{figure*}[h]
	\centering
	\subfloat [Instance segmentation in Entrance hall]{%
		\includegraphics[width=\columnwidth]{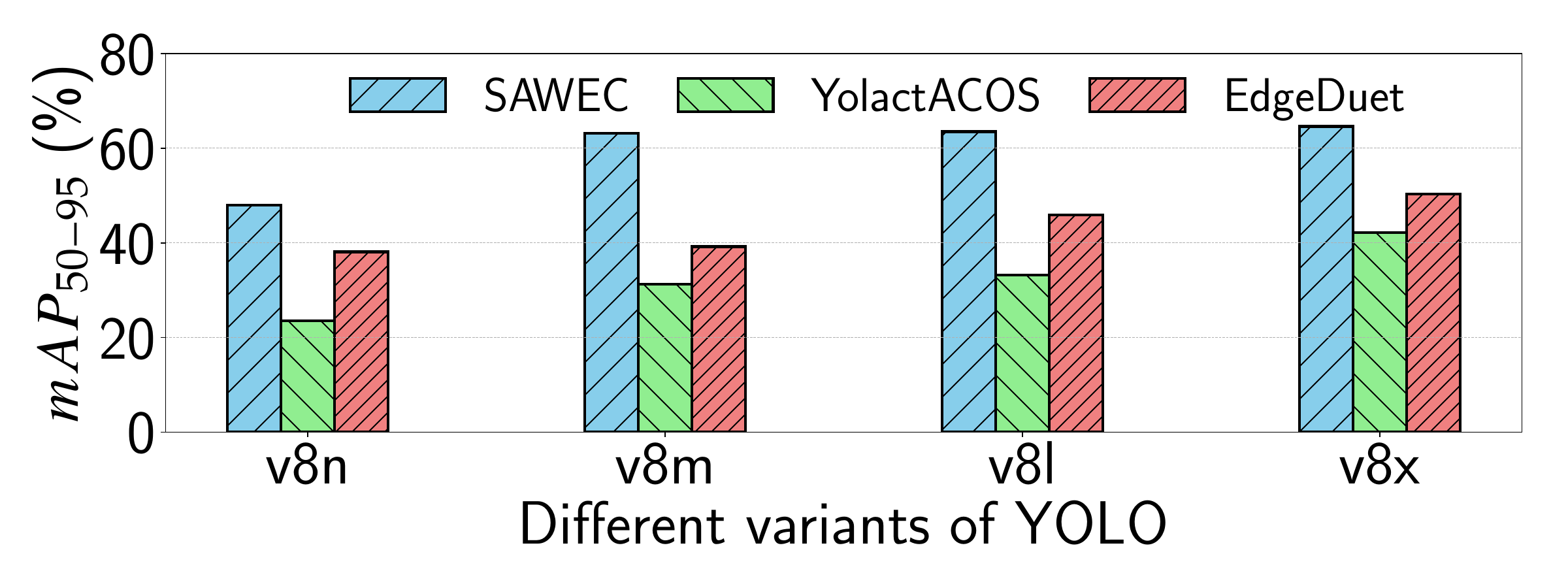}\label{fig:segmentation_performance_yolov8_entrance} }
	\hfill
		\subfloat[Instance segmentation in Anechoic chamber]{%
		\includegraphics[width=\columnwidth]{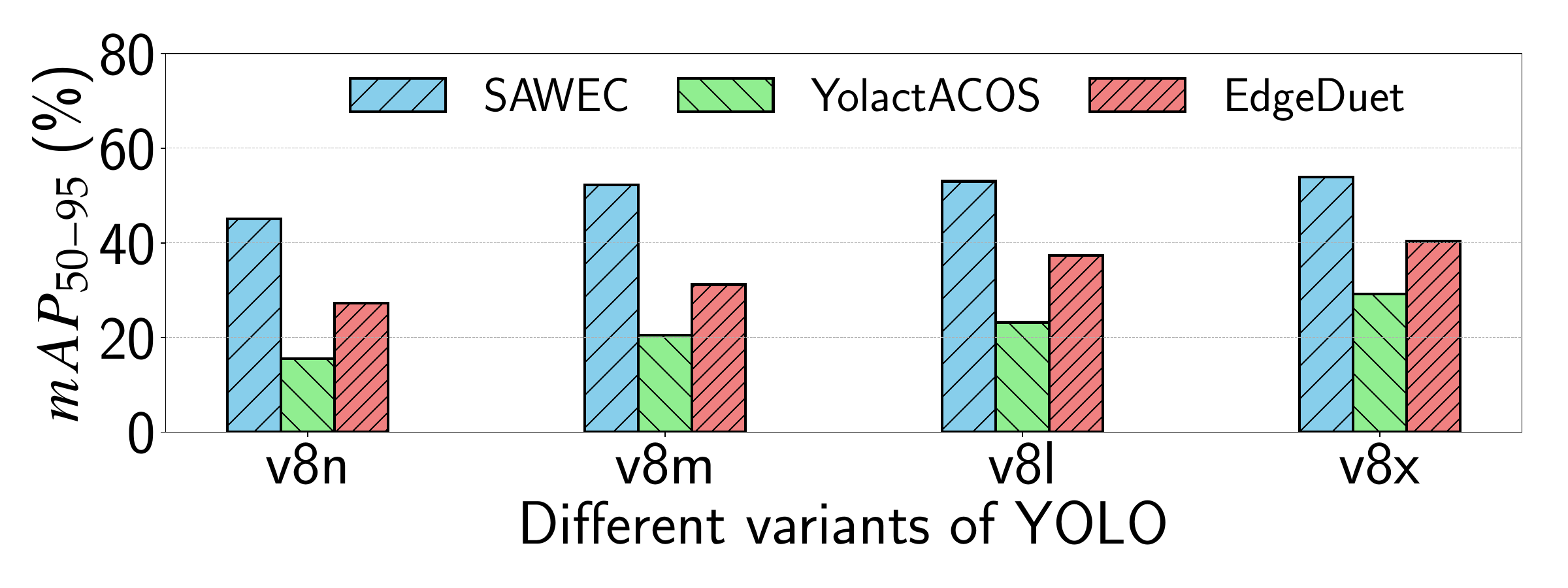}\label{fig:segmentation_performance_yolov8_anechoic}}
	\setlength\abovecaptionskip{0.1cm}
	\hfill
 	\subfloat [Object detection in Entrance hall]{%
		\includegraphics[width=\columnwidth]{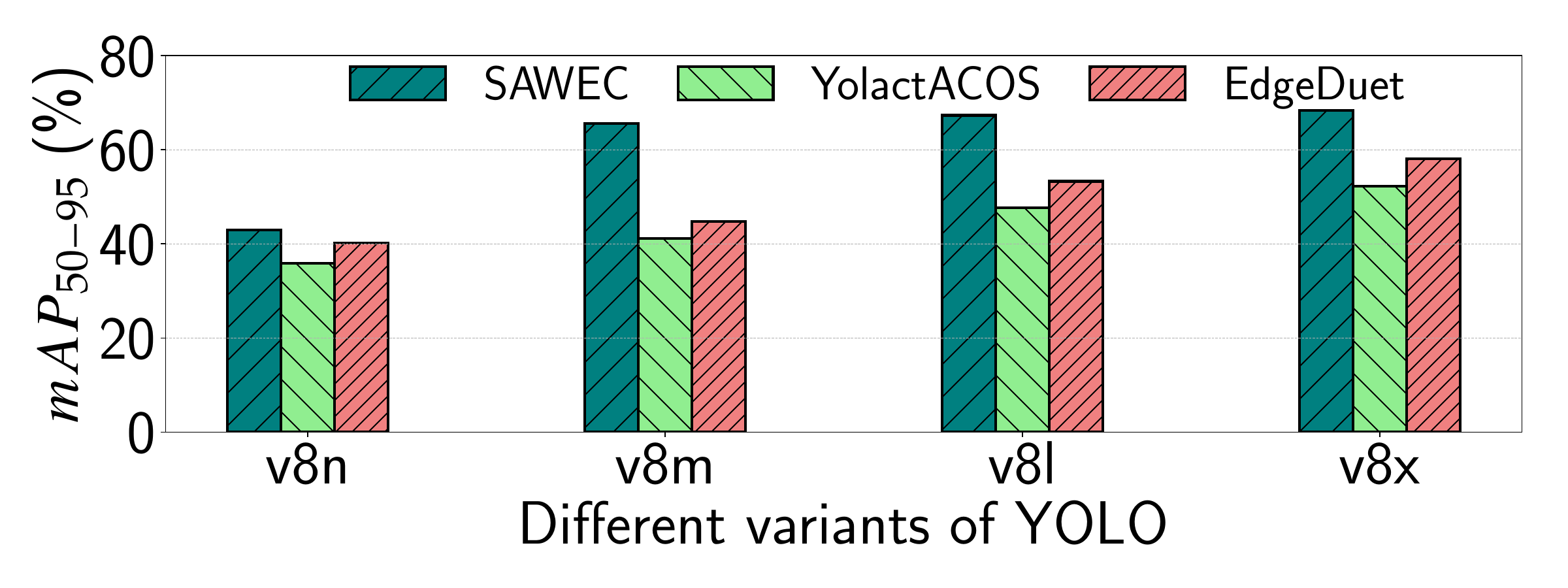}\label{fig:object_detection_performance_yolov8_entrance}}
	\hfill
		\subfloat[Object detection in Anechoic chamber]{%
		\includegraphics[width=\columnwidth]{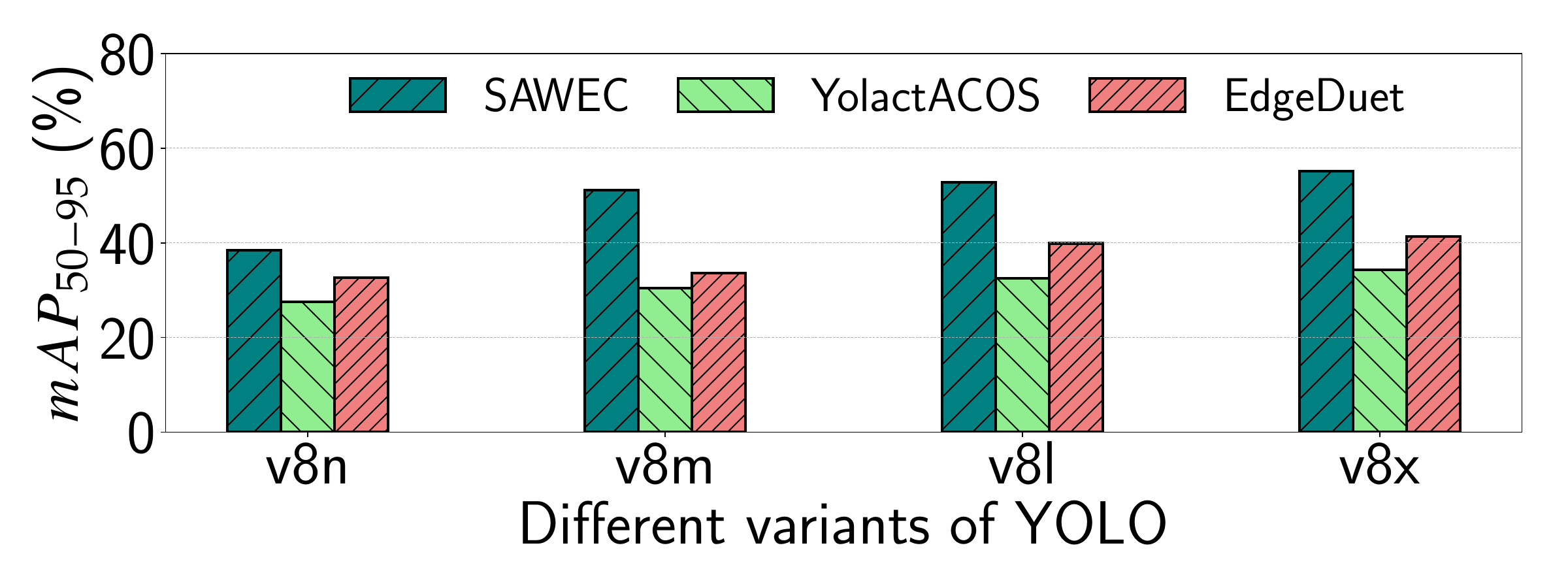}\label{fig:object_detection_performance_yolov8_anechoic}}
	\setlength\abovecaptionskip{0.1cm}
 
	\caption{$mAP_{50-95}$ with different variants of YOLOv8 for different edge computing approaches (\FW, YolactACOS, EdgeDuet).}\vspace{-0.1cm}
	\label{fig: segmentation_object_detection} 
\end{figure*}

The frames were transmitted from the camera to the edge server through a Wi-Fi link from an Intel AX200 \gls{nic} (2 antennas) at the camera and an Asus-RT-AX86u \gls{ap} (4 antennas) at the edge server.~The average link speed was $70.3$~MB/s. The edge server was connected to the \gls{ap} through an Ethernet link. The camera's \gls{nic} and the \gls{ap} were in the line of sight at a distance of $5$~m from each other.\vspace{-0.4cm}

\subsection{Performance for Different \gls{roi} Multiplying Factors $\alpha$}
In Figure \ref{fig:segmentation_performance_alpha} we present the instance segmentation and object detection performance of the YOLOv8m model \cite{Jocher_YOLO_by_Ultralytics_2023} when varying $\alpha$ from $1$ to $4$. The results show that with $\alpha=1$, $mAP_{50-95}$ is $28.8$\% and $32.2\%$ respectively for segmentation and object detection -- meaning that the performance is very poor. This is due to the fact that the \gls{roi} is too small to contain the pixels of the entire subject, as depicted in Figure \ref{fig:frame_translation}. Starting from $\alpha=2$ the \gls{roi} is sufficiently large to cover the subject (see Figure \ref{fig:frame_translation}) and, in turn, the $mAP_{50-95}$ reaches to $57.7$\% and $58.5\%$ respectively for segmentation and object detection. The performance slightly decreases when $\alpha\!=\!4$ as the image will contain more objects to be segmented and detected. On the other hand, the channel occupation keeps increasing when increasing $\alpha$ as this requires transmitting an increasing number of pixels. We have selected $\alpha\!=\!2$ for our evaluations as a tradeoff between computer vision task (instance segmentation and object detection) performance and channel occupation.

\begin{figure*}[h]
	\centering
	\subfloat[$mAP_{50-95}$ of image segmentation with original, resized, and compressed frames.]{
		\includegraphics[width=0.31\textwidth]{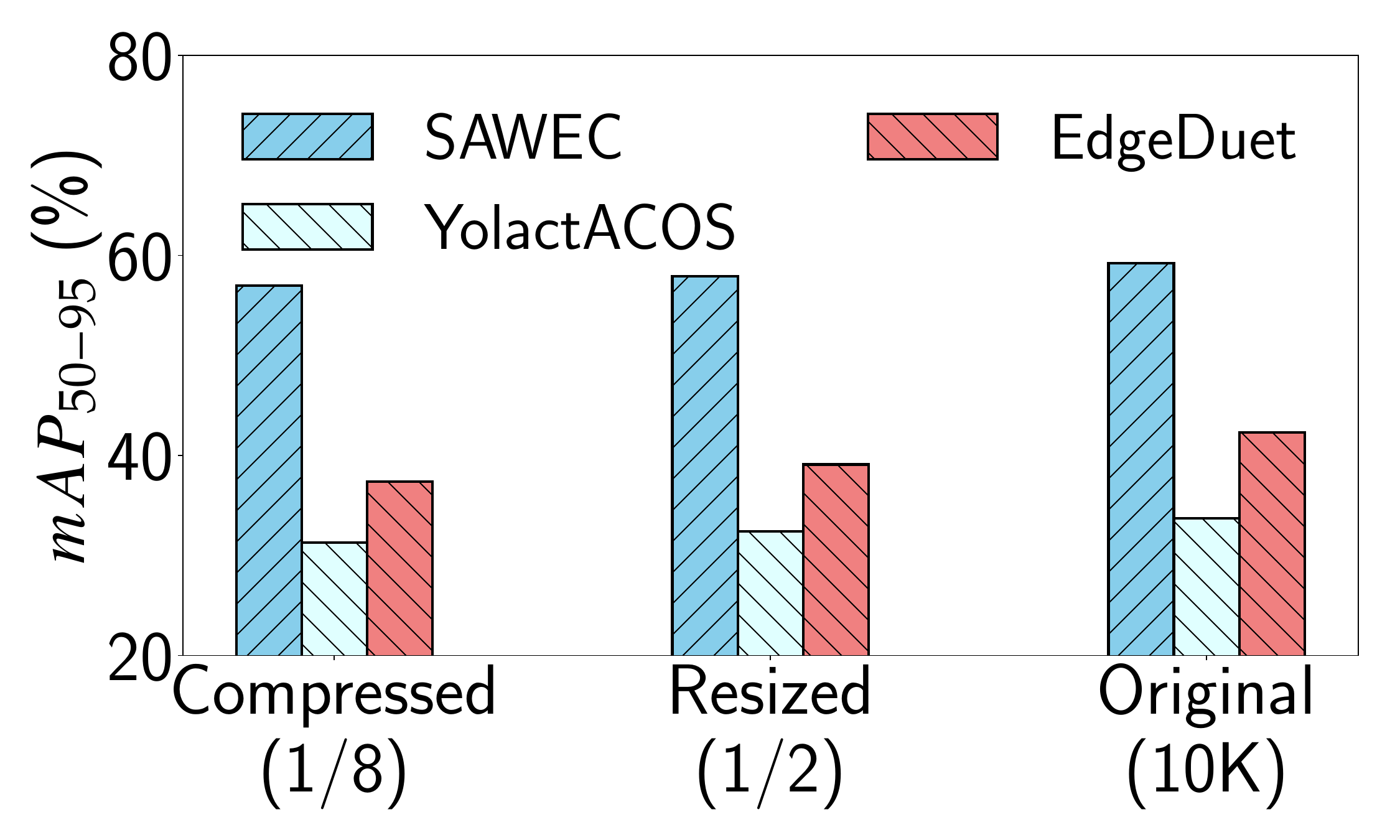}\label{fig:map_different_frame_shape}}
	\hfill
		\subfloat[Channel occupation of original, resized (1/2), and compressed (1/8) frames.]{%
		\includegraphics[width=0.33\textwidth]{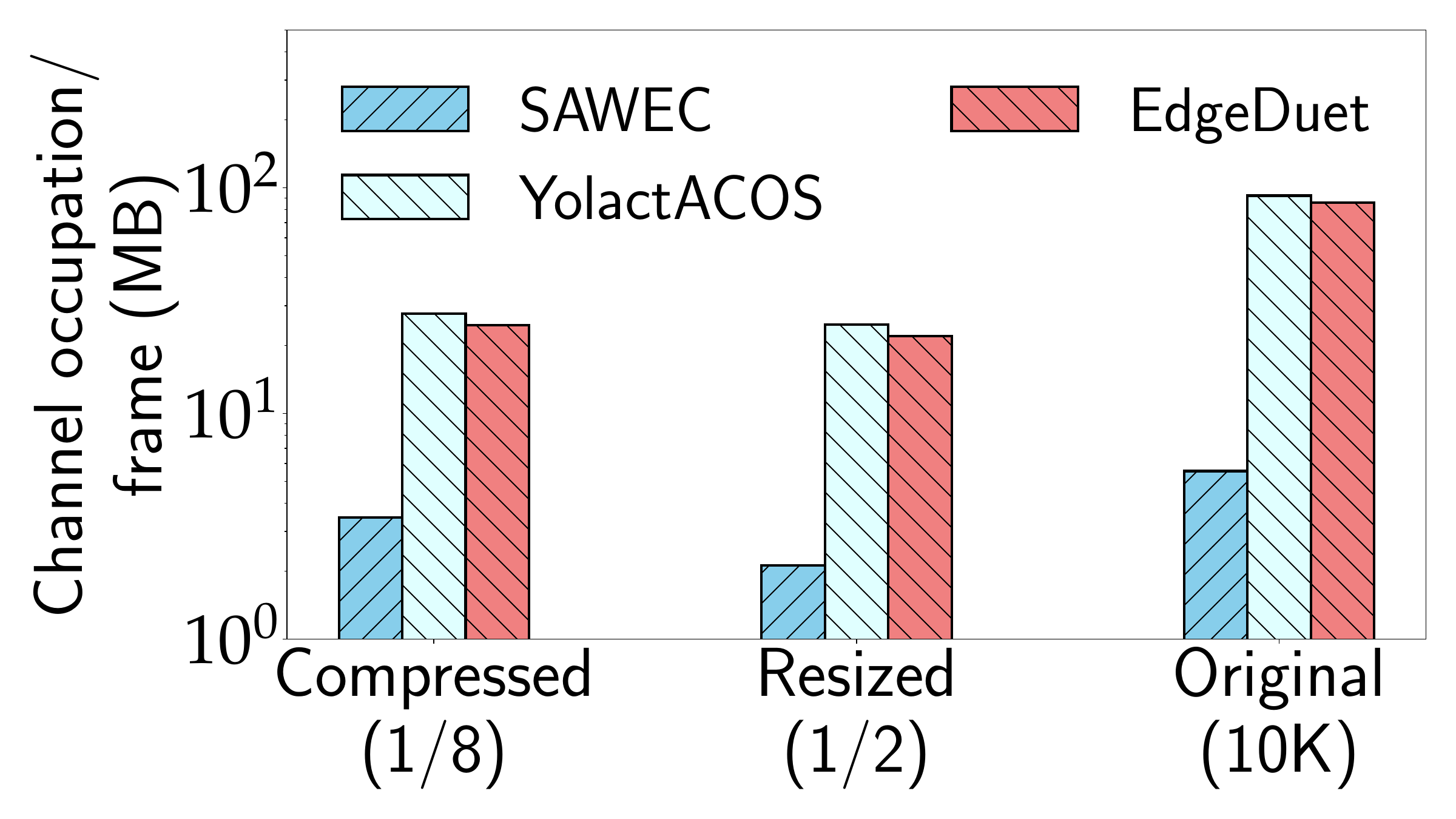}\label{fig:bandwidth_comparison} }
	\hfill
		\subfloat[End-to-end latency of original, resized (1/2), and compressed (1/8) frames.]{%
		\includegraphics[width=0.33\textwidth]{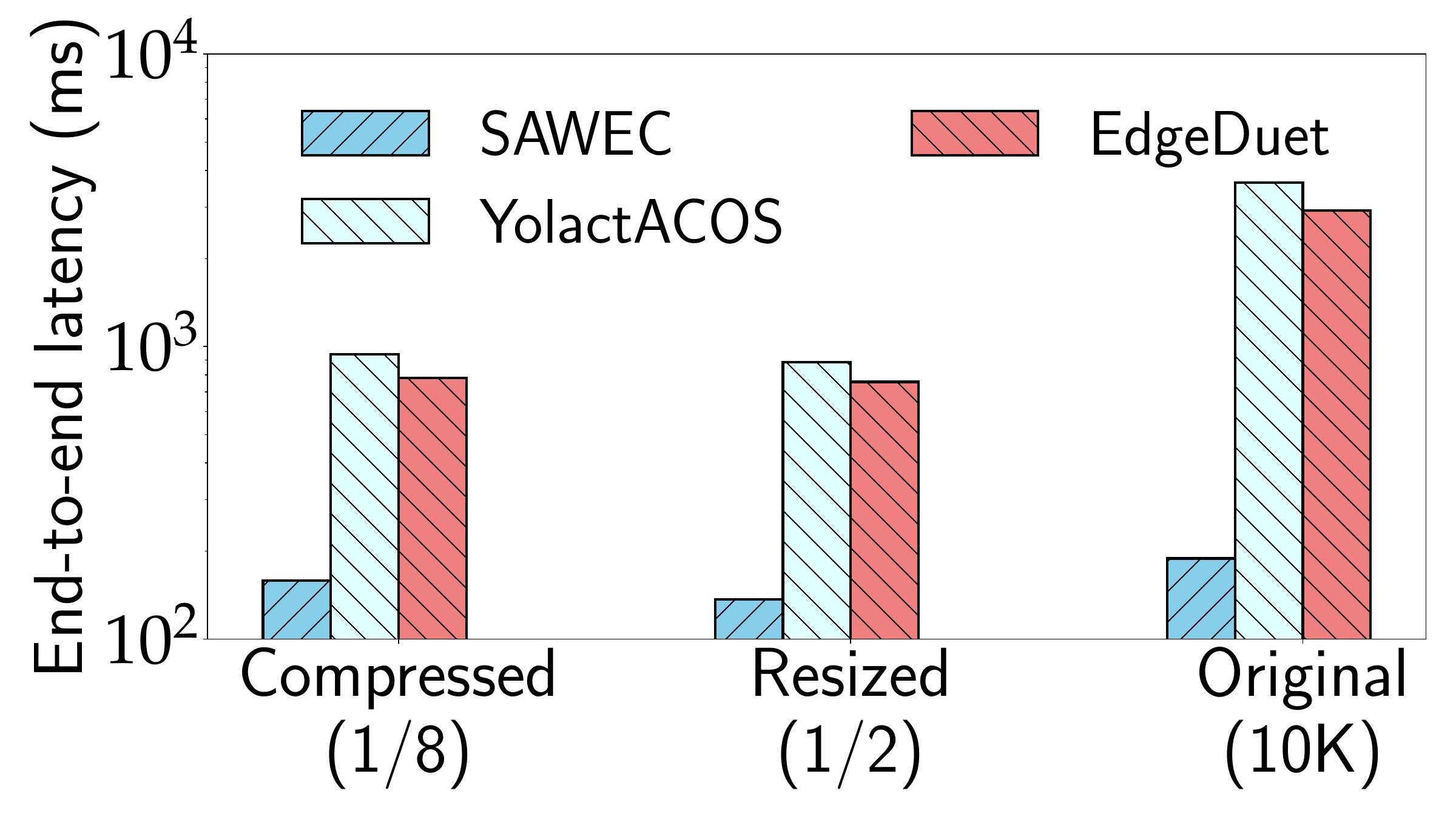}\label{fig:end-to-end_latency} }
	\caption{Comparative analysis of \FW with YolactACOS and EdgeDuet. The metrics are averaged over the two environments.\vspace{-0.2 cm}}
\end{figure*}

\subsection{Performance for Different Edge Computing Approaches}\label{sec:Segmentation Performance}

We assessed the image segmentation and object detection performance with the three different edge computing approaches mentioned before, i.e., \FW, YolactACOS, and EdgeDuet. We consider four different variants of \gls{sota} YOLOv8~\cite{Jocher_YOLO_by_Ultralytics_2023} model for our analysis: YOLOv8x, YOLOv8l, YOLOv8m, and YOLOv8n. 
According to the benchmark~\cite{Jocher_YOLO_by_Ultralytics_2023}, YOLOv8x is the highest-performing model with the highest inference latency whereas YOLOv8n is the lowest-performing model with the lowest inference latency. The performance of the instance segmentation task on original ($10$K) frames on the two evaluation environments -- entrance hall and anechoic chamber -- are presented respectively in Figures \ref{fig:segmentation_performance_yolov8_entrance}-\ref{fig:segmentation_performance_yolov8_anechoic}. 
Similar to the benchmark in \cite{Jocher_YOLO_by_Ultralytics_2023}, YOLOv8x performs best with all the approaches (\FW, YolactACOS, and EdgeDuet) whereas YOLOv8n performs the worst. Importantly, \FW 
achieves an average $mAP_{50-95}$ of $59.2\%$ over the two environments whereas YolactACOS and EdgeDuet achieved only $35.7\%$ and $45.3\%$ respectively on average. This is because with YolactACOS the $10$K image is downsized $1/16$ times by the model itself to fit the model input dimension of $640 \times 640$. This heavy downsizing hampers the performance severely. For the EdgeDuet, the performance is low as objects fall into multiple tiles which are processed separately, and small objects are not detected all the time. On the other hand, with \FW at $\alpha=2$, the average dimension of the detected \glspl{roi} is $996 \times 996$ which needs to be downsized by only $0.667$ times and it retains the higher quality of the images leading to a better performance.
The results for the object detection task follow the same trend and are reported in Figures \ref{fig:object_detection_performance_yolov8_entrance}-\ref{fig:object_detection_performance_yolov8_anechoic}. 

We also carried out a comparative performance analysis of \FW, YolactACOS, and EdgeDuet when the frames for all the approaches are downsized and compressed by $1/2$ and $1/8$ respectively (tolerable ratio as observed from the expression detection test in Section~\ref{sec: background_and_moti}). Figure \ref{fig:map_different_frame_shape} presents the $mAP_{50-95}$ of the instance segmentation performed using YOLOv8x with original ($10$K), resized ($1/2$), and compressed ($1/8$) versions of the video frames, considering the three different approaches for tasks offloading. The metric is the average of the results obtained in the entrance hall and the anechoic chamber evaluation environments. Note that similar results were obtained for object detection with original ($10$K), resized ($1/2$), and compressed ($1/8$) versions of the video frames. However, as the instance segmentation and object detection performance show similar trends -- as demonstrated in Figure \ref{fig: segmentation_object_detection} -- we only report the analysis for the image segmentation task. Similar to the observation of the preliminary facial expression detection test in Section~\ref{sec: background_and_moti}, the performance for all the approaches degrade slightly due to the image resizing and compression. Hence, a $1/2$ resize or~a $1/8$ compression can be used to reduce latency without hampering the image segmentation performance significantly. This might give \FW a further performance boost in terms of transmission latency and channel occupation. However, for a more delicate task like avatar and expression reconstruction, even downsizing and compressing by the ratio of only $1/2$ and $1/8$ respectively might affect the performance dramatically.\vspace{-0.2cm}

\subsection{Analysis of Channel Occupation}

Analyzing the per-frame channel occupation is paramount, considering that the radio spectrum is a pivotal and limited resource for wireless networks. 
The average per-frame channel occupation of \FW, YolactACOS, and EdgeDuet with three different frame types -- original (no resize and no compression), resized ($1/2$), and compressed ($1/8$) -- is depicted in Figure \ref{fig:bandwidth_comparison}. The results show that for original frames, \FW reduces the channel occupation by $94.03\%$ and $93.59\%$ respectively in comparison to YolactACOS and EdgeDuet. To elaborate further, on average, the size of an original frame at $10$K resolution is $138.88$ MB whereas the average channel occupation required by \FW, YolactACOS, and EdgeDuet are  $5.5$ MB, $92.2$ MB, and $85.9$ MB per frame respectively. Note that, the number of \gls{roi} is more than one for the tests with multiple persons moving simultaneously in the environment. In fact, the channel occupation of \FW might increase a bit depending on a higher number of moving subjects in the environments. However, to have an idea, even after having a total of $10$ \glspl{roi} (with $\alpha=2$) per single $10$K frame adds up to only $36$ MB reducing the channel occupation by $59.57\%$ on an average with respect to the discussed \gls{sota} approaches -- YolactACOS and EdgeDuet. \vspace{-0.1cm}

\begin{table}[]
\centering
\caption{Breakdown of end-to-end latency for original images using \FW, YolactACOS, and EdgeDuet.\vspace{-0.1cm}}
\label{tab:Breakdown of end-to-end latency}
\begin{tabular}{@{}cccc@{}}
\toprule
\multirow{2}{*}{\begin{tabular}[c]{@{}c@{}}Latency \\ components\end{tabular}} &
  \multicolumn{3}{c}{Approach} \\ \cmidrule(l){2-4} 
 &
  SAWEC &
  \begin{tabular}[c]{@{}c@{}}YolactACOS\end{tabular} &
  \begin{tabular}[c]{@{}c@{}}EdgeDuet\end{tabular} \\ \midrule
\begin{tabular}[c]{@{}c@{}}Tx time (ms)\end{tabular}          & \textbf{79.44}  & 1311    & 1221.62 \\ \cmidrule(r){1-1}
\begin{tabular}[c]{@{}c@{}}I/O time (ms)\end{tabular}         & \textbf{15.75}  & 528.46 & 16.73   \\ \cmidrule(r){1-1}
\begin{tabular}[c]{@{}c@{}}Inference time (ms)\end{tabular}   & \textbf{18.43}  & 19.79   & 18.56   \\ \cmidrule(r){1-1}
\begin{tabular}[c]{@{}c@{}}SAWEC proc. (ms) \end{tabular} & \textbf{75.0}   & 0       & 0       \\ \cmidrule(r){1-1}
\begin{tabular}[c]{@{}c@{}}End device latency (ms) \end{tabular} & \textbf{0}   & 1779.34       &  1667.18      \\ \cmidrule(r){1-1}
End-to-end (ms)                                                     & \textbf{188.62} & 3638.59 & 2920.91 \\ \bottomrule
\end{tabular}
\end{table}

\subsection{End-to-End Latency Analysis}
End-to-end latency is one of the most important key performance indicators for time-critical edge computing tasks including a wide range of \gls{vr} applications. We analyze the end-to-end latency of \FW by comparing it with YolactACOS, and  EdgeDuet for different frame types. The results are presented in Figure \ref{fig:end-to-end_latency}. The average end-to-end latency of \FW is $188.62$ ms, $136.67$ ms, and $158.46$ ms for original, resized ($1/2$), and compressed ($1/2$) frames respectively, which is much lower than the other two approaches. With $10$K resolution, \FW improves the latency by $94.80\%$ and $93.52\%$ in comparison to the YolactACOS and EdgeDuet respectively. To better understand, the end-to-end latency of the three approaches for the original 10K frames is broken down in Table \ref{tab:Breakdown of end-to-end latency}. We can see that the transmission (Tx) time and input/output (I/O) time for executing \FW are around $16$ times and $34$ times smaller than the corresponding values for YolactACOS. This is due to the higher channel occupation of the YolactACOS approach in comparison to \FW. On the other hand, I/O time for the EdgeDuet is similar to that of \FW as the whole frame is divided into smaller $640\times640$ tiles. However, this does not reduce the EdgeDuet Tx time which is around 15 times higher than \FW. 
Note that our benchmarking testbed has only one edge server. Deploying multiple edge servers will facilitate parallel offloading, thereby significantly reducing the transmission time across all three approaches.

Note that the effectiveness of \gls{roi} detection is contingent upon wireless localization and tracking. While \FW is well-suited for typical day-to-day scenarios, its performance is anticipated to diminish in highly dynamic environments. Moreover, determining the number of moving subjects that are tolerated by \FW before the performance degrades is a challenging problem, as it depends on (i) the distance among the subjects, (ii) the speed with which subjects are moving, (iii) the localization algorithm and (iv) the number of antennas that are used for localization. We believe these aspects deserve a separate investigation. Future research efforts will include improving the \FW performance in highly dynamic scenarios by enhancing the localization resolution.

\vspace{-0.1cm}
\section{Conclusions}
\vspace{-0.1cm}
In this paper, we proposed a new paradigm for \textit{sensing-assisted wireless edge computing}, in short, \FW. Our new approach leverages Wi-Fi-based localization and tracking to support high resource-consuming 360$^\circ$ computer vision tasks by obtaining the location of \gls{roi} based on the environment dynamics. This information allows offloading to the edge server only the detected \gls{roi} instead of the entire frame, thus reducing airtime overhead and overall latency. 
The \gls{aoa} and \gls{toa} estimated from the Wi-Fi \gls{cfr} are used to locate the \glspl{roi} in the 360$^\circ$ video frames. We assessed the effectiveness of \FW through extensive data collection campaigns in two different environments: an entrance hall and an anechoic chamber, each entailing six different test experiments. We prove the new concept by implementing a two-antenna IEEE 802.11ax-based Wi-Fi localization system. Our proposed approach reduces the overall end-to-end latency by $94.80\%$ and $93.52\%$ respectively while achieving $39.69\%$ and $23.64\%$ net $mAP_{50-95}$ improvement in comparison to the \gls{sota} \gls{wec} approach -- YolactACOS, and EdgeDuet -- in image segmentation tasks. Considering the object detection tasks, the average $mAP_{50-95}$ improvement is $42.72\%$ and $23.34\%$ respectively. Future research avenues include the localization performance improvement by increasing the number of antennas.
\vspace{-0.1cm}


\bibliographystyle{ACM-Reference-Format}
\bibliography{biblio,bib-francesco}

\end{document}

%% file: preamble.tex
\newcommand*{\FW}{\texorpdfstring{\texttt{SAWEC}}{SAWEC}\xspace}
\newcommand{\beq}{\begin{equation}}
\newcommand{\eeq}{\end{equation}}
\newcommand{\eq}[1]{Eq.~\eqref{#1}}

\newcommand{\cam}{360$^{\circ}$}

\usepackage{siunitx}
\usepackage{graphicx}
\usepackage{amsmath}
\usepackage{amsfonts}
\usepackage{optidef}
\usepackage{multirow}
\usepackage{xcolor}
\usepackage{caption}
\usepackage{subcaption}

\usepackage{hyperref}
\usepackage{comment}
\usepackage{lipsum}
\usepackage{graphicx}
\usepackage[utf8]{inputenc}
\usepackage{enumitem}
\usepackage{hyperref}
\usepackage{amsmath}
\usepackage{amsfonts}
\usepackage{bm}
\usepackage{tabularx}
\usepackage{svg}
\usepackage{algpseudocode}
\usepackage{fancyhdr}

\usepackage[linesnumbered,noend]{algorithm2e}

\AtBeginDocument{%
  \providecommand\BibTeX{{%
    \normalfont B\kern-0.5em{\scshape i\kern-0.25em b}\kern-0.8em\TeX}}}

\renewcommand\footnotetextcopyrightpermission[1]{} 
\usepackage{xspace}
\SetKwRepeat{Do}{do}{while}

\usepackage[acronyms,nonumberlist,nopostdot,nomain,nogroupskip,hyperfirst=false]{glossaries}

\input{tex/acronyms}
\RestyleAlgo{ruled}

\usepackage[many]{tcolorbox}
\usepackage{lipsum}

\definecolor{titlebg}{RGB}{100,22,72}
\definecolor{introbg}{RGB}{0,128,128}

\newtcolorbox{usecase}[1][]{
  breakable,
  enhanced,
  arc=0pt,
  outer arc=0pt,
  colframe=titlebg,
  colback=titlebg!05,
  overlay unbroken and first={
    \node[
      draw=titlebg,
      fill=titlebg,
      rotate=0,
      anchor=north west,
      text=white,
      font=\bfseries
    ]
    at (frame.north west)  
    {#1};
  }
}

\newtcolorbox{mission}[1][]{
  breakable,
  enhanced,
  arc=0pt,
  outer arc=0pt,
  colframe=introbg,
  colback=introbg!05,
  overlay unbroken and first={
    \node[
      draw=introbg,
      fill=introbg,
      rotate=0,
      anchor=north west,
      text=white,
      font=\bfseries
    ]
    at (frame.north west)  
    {#1};
  }
}

%% file: tex/acronyms.tex
\newacronym{6g}{6G}{sixth generation}
\newacronym{5g}{5G}{fifth generation}
\newacronym{snr}{SNR}{signal-to-noise ratio}
\newacronym{bw}{BW}{bandwidth}
\newacronym[plural=\gls{cnn}s,firstplural=convolutional neural networks (CNNs)]{cnn}{CNN}{convolutional neural network}

\newacronym{iq}{I/Q}{in phase/quadrature}
\newacronym{ml}{ML}{machine learning}
\newacronym{phy}{PHY}{physical}
\newacronym[plural=\gls{dnn}s,firstplural=deep neural networks (DNNs)]{dnn}{DNN}{deep neural network}
\newacronym{mmwave}{mmWave}{millimeter wave}
\newacronym{dsp}{DSP}{digital signal processing}
\newacronym{dsa}{DSA}{dynamic spectrum access}
\newacronym{ism}{ISM}{industrial, scientific and medical}
\newacronym{csi}{CSI}{channel state information}
\newacronym{fcc}{FCC}{Federal Communication Commission}
\newacronym{rfp}{RFP}{radio fingerprinting}
\newacronym[plural=\gls{sdr}s,firstplural=long training fields (SDRs)]{sdr}{SDR}{software-defined radio}

\newacronym{iot}{IoT}{Internet of things}
\newacronym{mimo}{MIMO}{multiple-input, multiple-output}
\newacronym{simo}{SIMO}{single-input, multi-output}
\newacronym{mum}{MU-MIMO}{multi-user MIMO}
\newacronym{sum}{SU-MIMO}{single-user multi-input, multi-output}
\newacronym{iui}{IUI}{inter-user interference}
\newacronym{isi}{ISI}{inter-stream interference}

\newacronym[plural=\gls{wlan}s,firstplural=wireless local area networks (WLANs)]{wlan}{WLAN}{wireless local area network}

\newacronym{ap}{AP}{access point}
\newacronym{sta}{STA}{station}
\newacronym{dl}{DL}{downlink}

\newacronym{mcs}{MCS}{modulation and coding scheme}
\newacronym{cfr}{CFR}{channel frequency response}
\newacronym{cir}{CIR}{channel impulse response}
\newacronym{ndp}{NDP}{null data packet}
\newacronym[plural=\gls{ltf}s,firstplural=long training fields (LTFs)]{ltf}{LTF}{long training field}
\newacronym{vht}{VHT}{very high throughput}
\newacronym{ht}{HT}{high throughput}

\newacronym{ofdm}{OFDM}{orthogonal frequency-division multiplexing}
\newacronym{ofdma}{OFDMA}{orthogonal frequency-division multiple access}
\newacronym{cfo}{CFO}{carrier frequency offset}
\newacronym{sfo}{SFO}{sampling frequency offset}
\newacronym{pdd}{PDD}{packet detection delay}
\newacronym{ppo}{PPO}{phase-locked loop offset}
\newacronym{pll}{PLL}{phase-locked loop}
\newacronym{pa}{PA}{phase ambiguity}
\newacronym{sbc}{SBC}{single board computer}

\newacronym[plural=\gls{cm}s,firstplural=confusion matrices (CMs)]{cm}{CM}{confusion matrix}

\newacronym{id}{ID}{identifier}
\newacronym{aoa}{AoA}{angle of arrival}
\newacronym{aod}{AoD}{angle of departure}
\newacronym{ul}{UL}{uplink}

\newacronym{svd}{SVD}{singular value decomposition}

\newacronym[plural=\gls{pdf}s,firstplural=probability density functions (PDFs)]{pdf}{PDF}{probability density function}

\newacronym{har}{HAR}{human activity recognition}  
\newacronym{cots}{COTS}{commercial-off-the-shelf} 
\newacronym[plural=\gls{imu}s,firstplural=inertial measurement units (IMUs)]{imu}{IMU}{inertial measurement unit}
\newacronym[plural=\gls{usrp}s,firstplural=universal software radio peripherals (USRPs)]{usrp}{USRP}{universal software radio peripheral}
\newacronym[plural=\gls{svm}s,firstplural=support vector machines (SVMs)]{svm}{SVM}{support vector machine}
\newacronym{ifft}{IFFT}{inverse fast Fourier transform}
\newacronym{fmcw}{FMCW}{frequency-modulated continuous-wave}
\newacronym[plural=\gls{enb}s,firstplural= evolved nodes base (eNBs)]{enb}{eNB}{evolved node base}

\newacronym{mac}{MAC}{medium access control}

\newacronym{he}{HE}{high-efficiency}
\newacronym{eht}{EHT}{extremely high throughput} 
\newacronym{ng60}{NG60}{next generation 60~GHz} 
\newacronym{3gpp}{3GPP}{Third Generation Partnership Project} 
\newacronym{nr}{NR}{new radio}
\newacronym{nru}{NR-U}{new radio-based access to unlicensed spectrum}
\newacronym{ieee}{IEEE}{Institute of Electrical and Electronics Engineers}
\newacronym{gsm}{GSM}{global system for mobile communications}
\newacronym{etsi}{ETSI}{European Telecommunications Standards Institute}
\newacronym{umts}{UMTS}{universal mobile telecommunications system}
\newacronym{lte}{LTE}{long term evolution of UMTS}
\newacronym{itu}{ITU}{International Telecommunications Union}
\newacronym{imt}{IMT}{International Mobile Telecommunications}
\newacronym{scfdma}{SC-FDMA}{single carrier - frequency division multiple access}
\newacronym{qos}{QoS}{quality of service}
\newacronym{qoe}{QoE}{quality of experience}

\newacronym{mmtc}{mMTC}{massive machine type communications}
\newacronym{urllc}{URLLC}{ultra-reliable and low latency communications}
\newacronym{embb}{eMBB}{enhanced mobile broadband}

\newacronym[plural=\gls{meh}s,firstplural=MEC hosts (MEHs)]{meh}{MEH}{MEC host}

\newacronym{gnss}{GNSS}{global navigation satellite system}

\newacronym[plural=\gls{mc}s,firstplural=Markov chains (MCs)]{mc}{MC}{Markov chain} 
\newacronym[plural=\gls{nn}s,firstplural=neural networks (NNs)]{nn}{NN}{neural network}
\newacronym[plural=\gls{rnn}s,firstplural=recurrent neural networks (RNNs)]{rnn}{RNN}{recurrent neural network}
\newacronym[plural=\gls{gru}s,firstplural=gated recurrent units (GRUs)]{gru}{GRU}{gated recurrent unit}

\newacronym{toa}{ToA}{time of arrival}
\newacronym{tdoa}{TDoA}{time difference of arrival}
\newacronym{rss}{RSS}{received signal strength}
\newacronym{rssi}{RSSI}{received signal strength indicator}
\newacronym{sumo}{SUMO}{simulation of urban mobility}
\newacronym{pl}{PL}{path loss} 
\newacronym{los}{LoS}{line of sight} 
\newacronym{nlos}{NLoS}{non line of sight} 
\newacronym{hdbscan}{HDBSCAN}{hierarchical density-based spatial clustering of applications with noise}

\newacronym{ue}{UE}{user equipment}
\newacronym{iov}{IoV}{Internet of vehicles}
\newacronym[plural=\gls{vm}s,firstplural=virtual machines (VMs)]{vm}{VM}{virtual machine}

\newacronym[plural=\gls{pv}s,firstplural=photovoltaic panels (PVs)]{pv}{PV}{photovoltaic panel}
\newacronym{mpc}{MPC}{model predictive control}
\newacronym{cpu}{CPU}{central processing unit}

\newacronym{qp}{QP}{quadratic programming}  
\newacronym{eh}{EH}{energy harvesting}
\newacronym{ups}{UPS}{uninterrupted power supply}
\newacronym{mcc}{MCC}{mobile cloud computing}
\newacronym{mip}{MIP}{mixed integer programming}
\newacronym{m2m}{M2M}{\mbox{machine-to-machine}}

\newacronym{wg}{WG}{Working Group}
\newacronym{tg}{TG}{Task Group}
\newacronym{par}{PAR}{Project Authorization Request}
\newacronym{frd}{FRD}{Functional Requirement Document}
\newacronym{bf}{TGbf}{IEEE 802.11bf}
\newacronym{sc}{SC}{Standards Committee}
\newacronym{ec}{EC}{Executive Committee}
\newacronym{sfd}{SFD}{Specification Framework Document}
\newacronym{ppdu}{PPDU}{\gls{phy} Protocol Data Unit}
\newacronym{sp}{S\&P}{security and privacy}
\newacronym{mmw}{mmWave}{millimeter wave}

\newacronym[plural=\gls{nap}s,firstplural=non-AP stations (non-AP STAs)]{nap}{non-AP STA}{non-AP station}

\newacronym[plural=\gls{ru}s,firstplural=resource units (RUs)]{ru}{RU}{resource unit}
\newacronym{sage}{SAGE}{space-alternating generalized expectation maximization}
\newacronym{omp}{OMP}{orthogonal matching pursuit}
\newacronym{iht}{IHT}{iterative hard thresholding}
\newacronym{ar}{AR}{augmented reality}
\newacronym{vr}{VR}{virtual reality}
\newacronym{nic}{NIC}{network interface card}
\newacronym{tof}{ToF}{Time-of-Flight}
\newacronym{2d aoa}{2D AoA}{two dimensional angle of arrival}
\newacronym{wec}{WEC}{wireless edge computing}
\newacronym{sota}{SOTA}{state-of-the-art}
\newacronym{roi}{ROI}{region of interest}
\newacronym{mAP}{mAP}{mean average precision}
\newacronym{i/o}{I/O}{input-output}
\newacronym{fps}{FPS}{frames per second}
\newacronym{dbscan}{DBSCAN}{density-based spatial clustering of applications with noise}
\newacronym{ram}{RAM}{random access memory}
\newacronym{gpu}{GPU}{graphics processing unit}
\newacronym{flops}{FLOPs}{floating point operations per second}
\newacronym{iou}{IoU}{intersection over union}